\documentclass{article}

\usepackage{iclr2027_conference,times}
\iclrfinalcopy 

\usepackage{latexsym}
\usepackage[T1]{fontenc}
\usepackage[utf8]{inputenc}
\usepackage{microtype}
\usepackage{inconsolata}

\usepackage{url}
\providecommand{\href}[2]{#2}
\usepackage{booktabs}
\usepackage{graphicx}
\usepackage{wrapfig}
\usepackage{float}
\usepackage{subcaption}
\usepackage{multirow}
\usepackage{amsmath}
\usepackage{amsfonts}
\usepackage{amssymb}
\usepackage{bm}
\usepackage{nicefrac}
\usepackage{xcolor}
\usepackage{tcolorbox}
\usepackage{array}
\usepackage{enumitem}

\newcommand{\vtheta}{\bm{\theta}}
\newcommand{\vdelta}{\bm{\delta}}
\newcommand{\ve}{\bm{e}}
\newcommand{\vg}{\bm{g}}
\newcommand{\vu}{\bm{u}}
\newcommand{\vd}{\bm{d}}
\newcommand{\vr}{\bm{r}}

\newcommand{\mH}{\bm{H}}
\newcommand{\mP}{\bm{P}}
\newcommand{\mI}{\bm{I}}
\newcommand{\mW}{\bm{W}}
\newcommand{\vv}{\bm{v}}
\newtcolorbox{summarybox}{
  colback=blue!5,
  colframe=blue!60!black,
  fonttitle=\bfseries,
  arc=2mm,
  boxrule=0.8pt,
  left=2mm,
  right=2mm,
  top=1mm,
  bottom=1mm
}

\title{Local Conflict, Targeted Repair: A Perturbation Analysis of Multi-Domain RL}

\author{%
  Lei Yang\textsuperscript{1} \quad Siyu Ding\textsuperscript{2}\thanks{Project leader}  \quad Deyi Xiong\textsuperscript{1}\thanks{Corresponding author} \\
  \textsuperscript{1}TJUNLP Lab, College of Intelligence and Computing, Tianjin University, Tianjin, China \\
  \textsuperscript{2}Baidu Inc., Beijing, China \\
  \texttt{yanglei\_9@tju.edu.cn}
}

\begin{document}

\maketitle

\lhead{Preprint. Under review.}   


\begin{abstract}
Reinforcement learning (RL) post-training improves large language models on individual domains but can degrade previously acquired capabilities. We study this interference in mathematical reasoning, code generation, question answering (QA), and creative writing (CW). Full-model domain gradients can be nearly orthogonal despite substantial selective degradation. Our structural analyses show sparse, weakly overlapping edits on computation routes that reasoning domains share, with directional alignment associated with transfer and conflict. A local perturbation model separates loss changes into a first-order term and a path-averaged curvature term, decomposes the latter along the actual rollback intervention, and provides a conditional explanation for recovery by short refresh. The experiments support this local sensitivity account. After Code $\rightarrow$ Math $\rightarrow$ QA $\rightarrow$ CW training, a short Re-Math refresh raises Math from 57.66 to 66.04 and achieves the highest average score among the evaluated baselines. A training-free rollback at a $9.8\%$ budget recovers $57.4\%$ of the QA-induced Math accuracy drop with a small QA change. Curvature probes at the Math checkpoint further show higher sensitivity along the selected direction than along matched random directions. Together, these results support sparse intervention and targeted refresh as effective ways to mitigate cross-domain degradation.

\end{abstract}


\section{Introduction}



RL has become a primary technique for improving large language models (LLMs) across reasoning, coding, question answering, and open-ended generation. Real deployments need a single model strong across all of these domains rather than a collection of single-domain experts, and model iteration makes this demand incremental: new domain capabilities are often added on top of an already-trained model. The two standard recipes are joint (mixed) training over all domains at once and sequential training one domain after another.

\begin{wraptable}{r}{0.48\textwidth}
\centering
\small
\vspace{-1.5em}
\caption{Motivating example under the Omni-Thinker curriculum: RL improves the current domain but later stages selectively damage Math.}
\label{tab:intro_motivation}
\resizebox{\linewidth}{!}{
\begin{tabular}{l|cccc}
\toprule
 & $\mathrm{Code}_o$ & $\mathrm{Math}_o$ & $\mathrm{QA}_o$ & $\mathrm{CW}_o$ \\
\midrule
Code & \textbf{52.67} & 50.69 & 50.99 & 50.47 \\
Math & \textcolor{gray!45}{59.63} & \textbf{66.49} & \textcolor{red}{59.90} & \textcolor{red}{57.66} \\
QA   & \textcolor{gray!45}{60.89} & \textcolor{gray!45}{60.52} & \textbf{62.34} & 62.34 \\
CW   & \textcolor{gray!45}{82.40} & \textcolor{gray!45}{81.44} & \textcolor{gray!45}{81.79} & \textbf{86.52} \\
\bottomrule
\end{tabular}
}
\end{wraptable}

A sequential curriculum makes the problem clear. Under the Omni-Thinker~\citep{li2026omnithinker} curriculum Code $\rightarrow$ Math $\rightarrow$ QA $\rightarrow$ CW (Table~\ref{tab:intro_motivation}), Math rises to 66.49 after its own stage but falls to 57.66 after the later QA and CW stages, while Code and QA remain comparatively stable. In our comparisons, joint training and replaying all four domains also leave Math below its single-domain expert (Section~\ref{sec:exp_refresh}). These observations motivate a closer look at the mechanism of forgetting~\citep{kirkpatrick2017overcoming,shi2024continual}: the full-model gradient cosine stays near zero even as an earlier domain is damaged, so global gradient conflict~\citep{sener2018multi,liang2026cgpo} alone does not characterize the observed interference. Where does the relevant cross-domain sensitivity reside?

We examine the local routes through which updates act. Layer- and module-level analysis reveals localized conflict and synergy beneath global near-orthogonality. Single-domain RL produces sparse, small-magnitude edits with weak overlap among the top-changed neurons. Meanwhile, reasoning domains share many active computation routes, and the direction of their edits on these routes is associated with transfer and conflict. These observations motivate a local account of interference beyond global gradient alignment or edit overlap alone.

We formalize this account with a local perturbation model. The loss change caused by a later-domain displacement has an exact decomposition into a signed first-order term and a path-averaged curvature term. Splitting the displacement into the part removed by a sparse intervention and the part retained identifies the corresponding curvature contributions and their interaction. Under local positive-curvature and weak-coupling conditions, an idealized short refresh has a geometric upper bound on its harmful displacement. This theory offers a conditional explanation for selective degradation and repair; the empirical case rests on the intervention and recovery results.

We examine the account through three experiments. First, a short Re-Math refresh from $\mathrm{CW}_o$ raises Math from 57.66 to 66.04, close to its single-domain expert, with limited changes to the other domains and the highest average score among the evaluated baselines (66.39). Second, training-free rollback recovers $20.4\%$ of the QA-induced Math accuracy drop at a $1.3\%$ budget and $57.4\%$ at $9.8\%$, with small QA changes. Third, at $\mathrm{Math}_o$, Math curvature along the selected direction is $4$--$7\times$ that of matched random directions. Together, these results support sparse intervention and targeted refresh as repair strategies, with directional curvature providing a complementary view of local sensitivity.


Our contributions are threefold:
\begin{itemize}[label=\textbullet,align=left,labelindent=0pt,labelwidth=0.6em,labelsep=0.45em,leftmargin=1.05em,topsep=3pt,itemsep=2pt,parsep=0pt,partopsep=0pt]
    \item We identify local structure in cross-domain interference through sparse edits, shared active routes, and directional alignment, despite weak full-model gradient conflict.
    \item We derive an exact local loss decomposition using the actual rollback operator, including fractional reversion of shared key and value rows, and establish a conditional bound for short refresh.
    \item We demonstrate effective recovery through short targeted refresh and sparse training-free rollback, supported by directional curvature diagnostics.
\end{itemize}

\section{Related Work}

RL post-training and multi-domain RL mainly optimize aggregate performance, leaving cross-domain interactions inside the model poorly understood. The line runs from preference-based RLHF~\citep{DBLP:conf/nips/ChristianoLBMLA17,DBLP:conf/nips/StiennonO0ZLVRA20,schulman2017proximal} through process supervision and large-scale reasoning RL~\citep{DBLP:conf/iclr/LightmanKBEBLLS24,shao2024deepseekmath,guo2025deepseekr1,team2025kimi} to recent multi-domain methods such as Omni-Thinker~\citep{li2026omnithinker} and CGPO~\citep{liang2026cgpo}. We instead ask where cross-domain interference resides, why it is selective, and why a short domain refresh restores performance.

Cross-domain degradation is often discussed in terms of continual learning (CL), forgetting, or gradient conflict. Recent studies show that apparent forgetting can reflect shifts in alignment or policy behavior~\citep{zheng2025spurious}. SFT--RL comparisons likewise suggest that RL may yield milder drift~\citep{chu2025sft,lai2025rft,shenfeld2025rlrazor,chen2025retaining}, with additional evidence for reasoning transfer~\citep{huan2025mathtransfer}. Multi-task optimization offers a complementary global view through gradient balancing~\citep{DBLP:conf/icml/ChenBLR18,sener2018multi} and gradient surgery~\citep{yu2020gradient,DBLP:conf/nips/LiuLJSL21}. Our results suggest that full-model gradients can appear nearly orthogonal even when localized conflict harms earlier domains.

Task-vector and mechanistic work suggests that model changes can often be localized. In weight space, model soups~\citep{wortsman2022model} and TIES-style merging~\citep{yadav2023ties} attribute transfer and interference to redundant or sign-conflicting updates~\citep{DBLP:conf/nips/MatenaR22}. Mechanistic studies localize knowledge and skills to compact subsets of parameters~\citep{DBLP:conf/iclr/BauBSDDG19} or neurons~\citep{dai2022knowledge,wang2022finding,leng2024towards}. Recent feature-level analyses suggest that RL mostly preserves the base representation while strengthening compact task-relevant features~\citep{shi2026why}. We build on this locality view to study changed neurons, active routes, and direction-dependent interference.

\begin{figure*}[!t]
  \centering
  \begin{subfigure}[t]{0.32\textwidth}
    \centering
    \includegraphics[width=\linewidth]{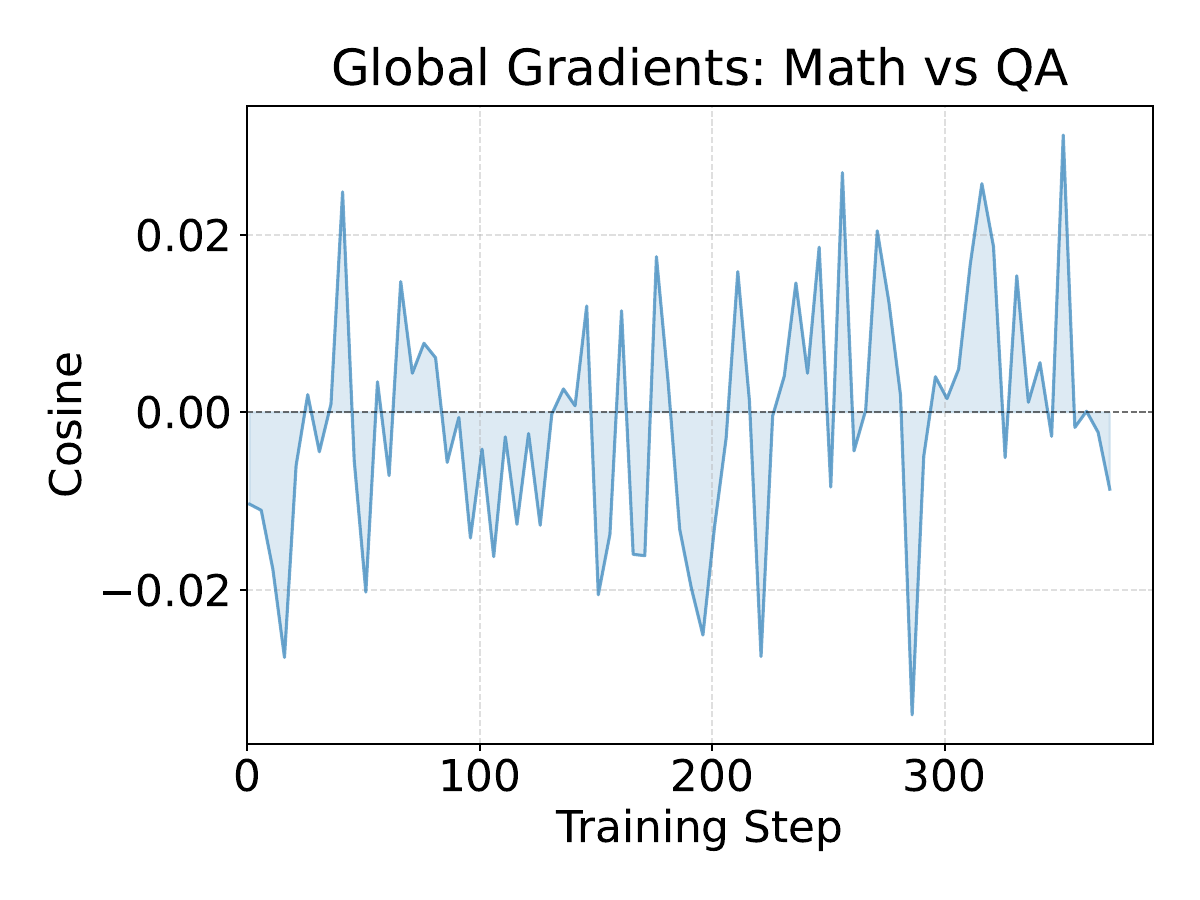}
    \caption{Full-model gradient cosine.}
    \label{fig:grad_conflict_cosine}
  \end{subfigure}\hfill
  \begin{subfigure}[t]{0.32\textwidth}
    \centering
    \includegraphics[width=\linewidth]{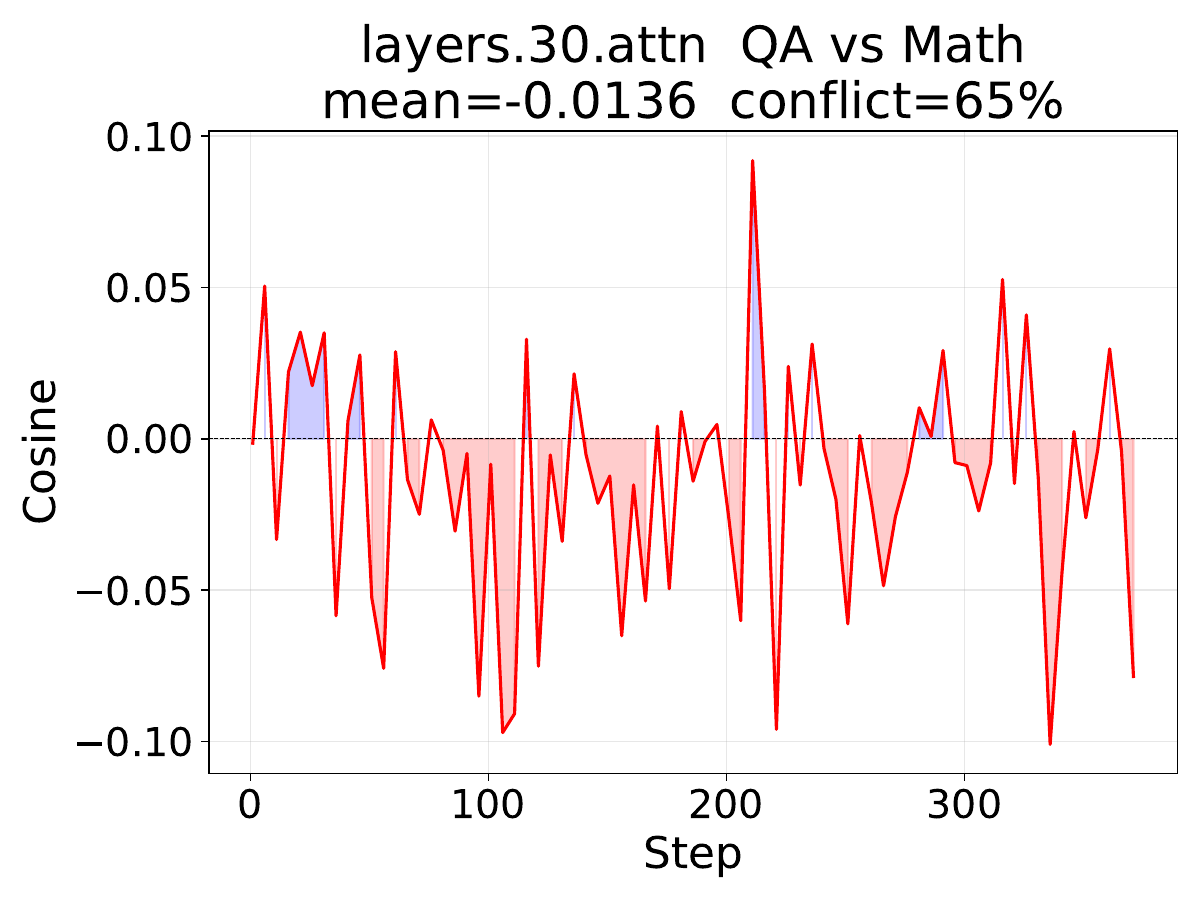}
    \caption{Most conflicting module.}
    \label{fig:attn_mlp_conflict}
  \end{subfigure}\hfill
  \begin{subfigure}[t]{0.32\textwidth}
    \centering
    \includegraphics[width=\linewidth]{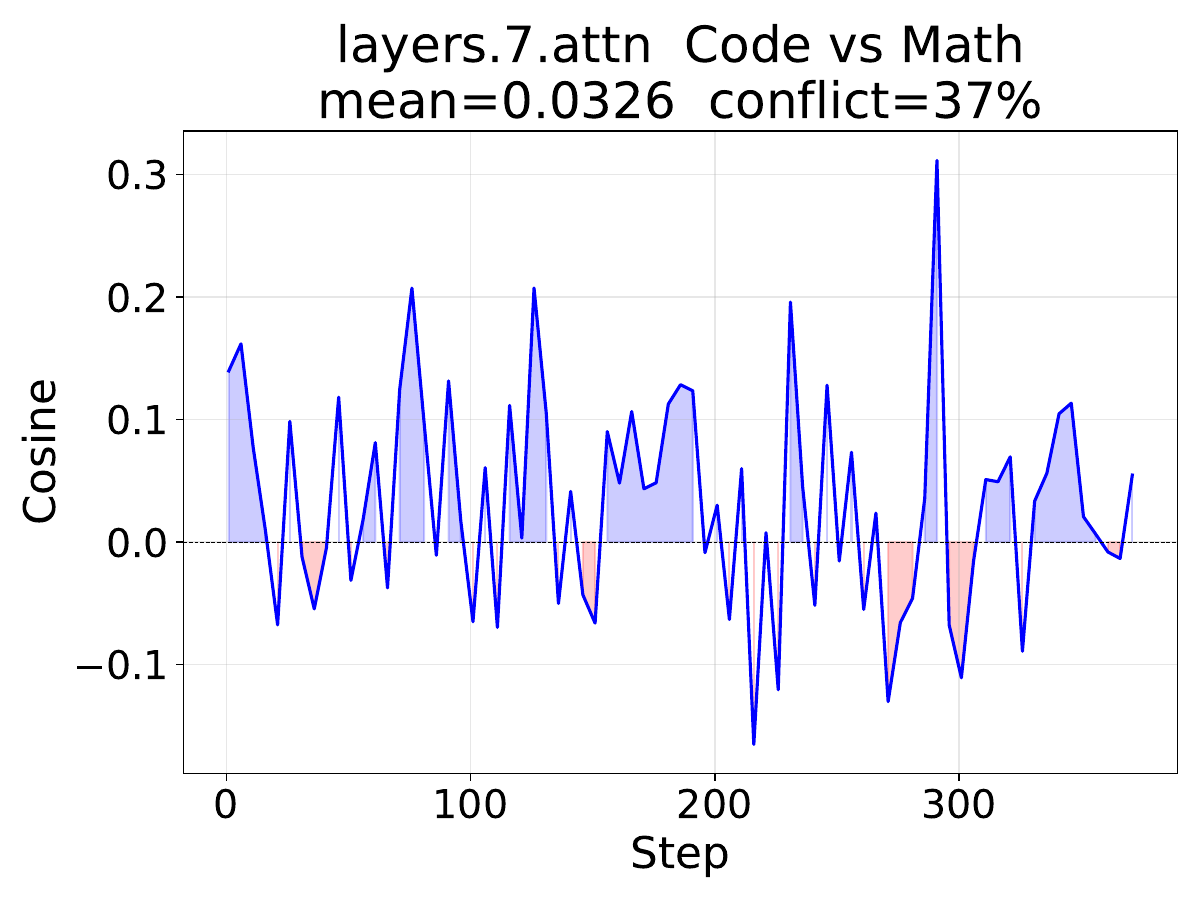}
    \caption{Most synergistic module.}
    \label{fig:attn_mlp_synergy}
  \end{subfigure}
  \caption{Gradient-level view of Math-QA interference in joint training: the full-model cosine stays near zero (a), yet a per-module decomposition reveals consistent conflict (b) and synergy (c).}
  \label{fig:grad_conflict}
\end{figure*}

\section{Empirical Study Setup}
\label{sec:setup}

To investigate where cross-domain interference arises and how recovery later occurs, we construct a controlled multi-domain RL setting on Qwen3-4B-Thinking-2507~\citep{DBLP:journals/corr/abs-2505-09388} over four domains: mathematical reasoning (Math), code generation (Code), question answering (QA), and creative writing (CW).



\textbf{Data.} Math is sampled from OpenR1-math~\citep{openr1}, Code from KlearReasoner-CodeSub-15K~\citep{su2025cegppocontrollingentropygradientpreserving}, QA from SuperGPQA~\citep{DBLP:journals/corr/abs-2502-14739} stratified by subfield and difficulty, and CW from the crownelius/Creative-Writing series\footnote{\url{https://huggingface.co/Crownelius}} (details in Appendix~\ref{app:data_const}).



\begin{wrapfigure}{r}{0.55\textwidth}
  \vspace{-1.2em}
  \centering
  \includegraphics[width=\linewidth]{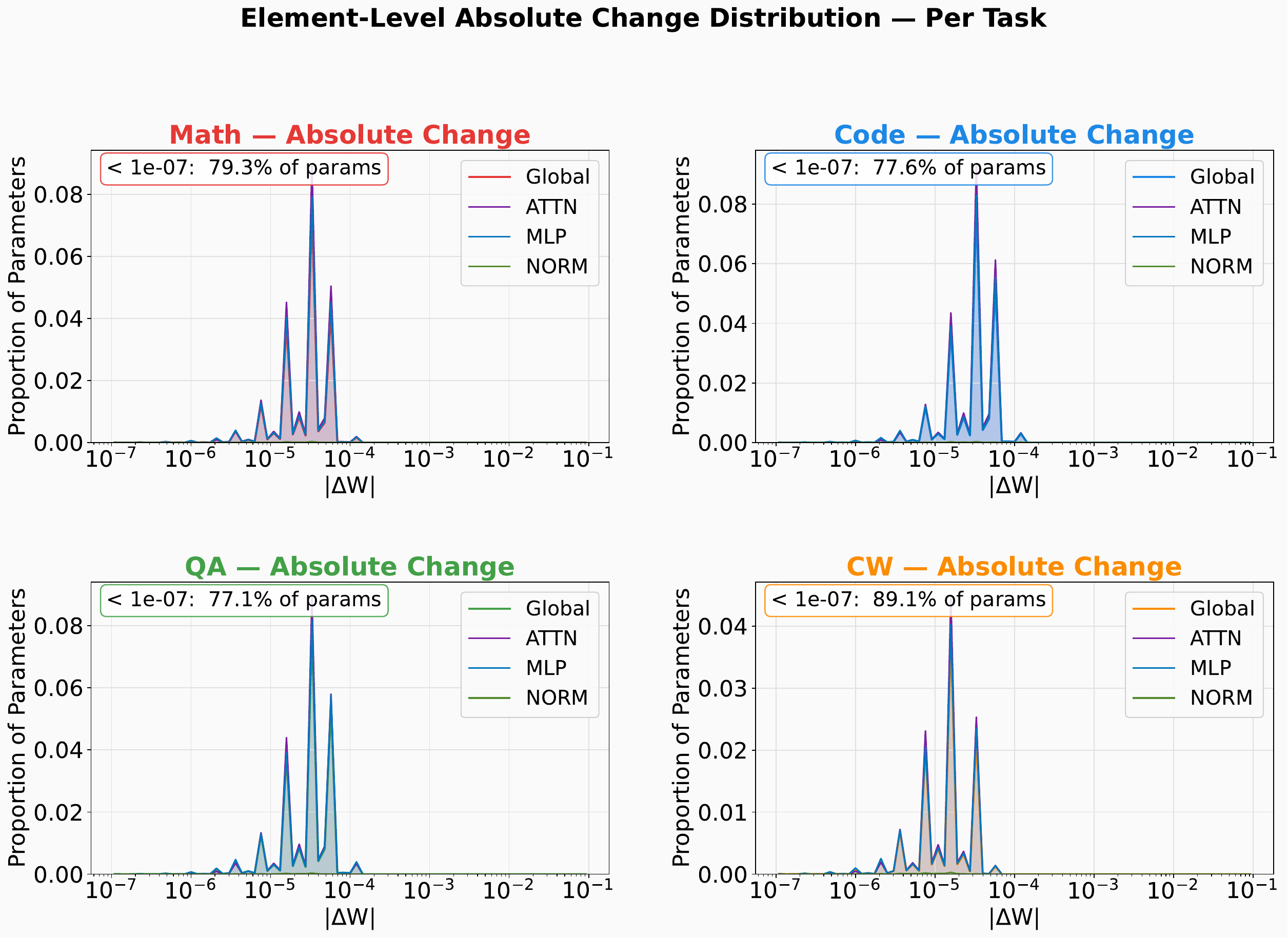}
  \caption{Absolute parameter changes of the four single-domain experts relative to the base model.}
  \label{fig:param_abs_per_task}
  \vspace{-1.0em}
\end{wrapfigure}

\textbf{Training.} All runs use GRPO~\citep{shao2024deepseekmath} in VeRL~\citep{DBLP:conf/eurosys/ShengZYWZZPL025} with per-domain reward functions (Appendix~\ref{app:reward_func}); except for the training domain and the initialization checkpoint, all experiments share the hyperparameters of Appendix~\ref{app:train_hyper}. We first train four single-domain experts from the base model ($\mathrm{Math}_s$, $\mathrm{Code}_s$, $\mathrm{QA}_s$, $\mathrm{CW}_s$), then run sequential multi-domain training in the fixed Omni-Thinker order~\citep{li2026omnithinker} Code $\rightarrow$ Math $\rightarrow$ QA $\rightarrow$ CW, each stage continuing from the previous checkpoint and yielding $\mathrm{Code}_o$, $\mathrm{Math}_o$, $\mathrm{QA}_o$, and $\mathrm{CW}_o$. Section~\ref{sec:exp} adds two four-domain mixed baselines: $\mathrm{JT}$, naive joint training in which each batch contains equal amounts of data from all domains and parameters are updated in the standard way, and $\mathrm{CGPO}$~\citep{liang2026cgpo}, which follows the official implementation in computing domain-wise updates within each batch and then updating the affected parameters with a specific learning rate.



\textbf{Checkpoints and evaluation.} Each stage trains to convergence and selects the checkpoint with the best validation performance on its current training domain. Final evaluation uses independent benchmarks: AIME24/25/26~\citep{aops_aime}, OlympiadBench~\citep{DBLP:conf/acl/HeLBHTSHHHZLQL024} and HMMT~\citep{dekoninck2026matharena} for Math; LiveCodeBench-v6~\citep{DBLP:conf/iclr/JainHGLYZWSSS25} for Code; SuperGPQA-test~\citep{DBLP:journals/corr/abs-2502-14739} and MMLU-Pro~\citep{DBLP:conf/nips/WangMZNCGRAHJLK24} for QA; and WritingBench~\citep{DBLP:journals/corr/abs-2503-05244} for CW (Appendix~\ref{app:data_const}).


\section{Structural Evidence for Localized Cross-Domain Interference}
\label{sec:ana}

This section asks where cross-domain interference resides. We proceed by elimination: full-model gradients rule out global antagonism, and sparse, weakly overlapping parameter edits rule out co-editing of the same weights, which shifts the perspective from training to the more direct inference stage, taking shared active routes as the candidate site. We then show that the direction of the edits carried on those routes is what separates transfer from damage.

\subsection{Global Gradients Miss Local Conflict}
\label{subsec:global_grad}

A natural first question is whether cross-domain interference in domain RL can be explained by global gradient conflict. During $\mathrm{JT}$ training, domain-specific gradients are periodically computed and the cosine similarity between domain pairs is measured, $\cos(\vg_{d_i}, \vg_{d_j}) = \frac{\vg_{d_i}^\top \vg_{d_j}}{\|\vg_{d_i}\|_2\,\|\vg_{d_j}\|_2}$, where $\vg_d$ is the gradient for domain $d$. Positive, negative, and near-zero values indicate alignment, conflict, and approximate orthogonality, respectively.


Figure~\ref{fig:grad_conflict_cosine} shows that the full-model gradient cosine between Math and QA stays close to zero throughout training, so the later drop in Math cannot be fully attributed to full-model gradient antagonism. Global near-orthogonality does not, however, imply that interference is absent. A coarse first probe is to decompose the same gradients by layer and module: individual attention and MLP modules then show consistent conflict (Figure~\ref{fig:attn_mlp_conflict}) or synergy (Figure~\ref{fig:attn_mlp_synergy}), most visibly for the Math-QA pair (Appendix~\ref{app:grad_conf}). Because this decomposition is at whole-module granularity and describes instantaneous gradients, we treat it only as a hint that interference is localized.





\subsection{Edits Are Sparse and Small}
\label{subsec:sparse_update}

\begin{wrapfigure}{r}{0.37\textwidth}
  \vspace{-2.5em}
  \centering
  \begin{subfigure}[t]{\linewidth}
    \centering
    \includegraphics[width=\linewidth]{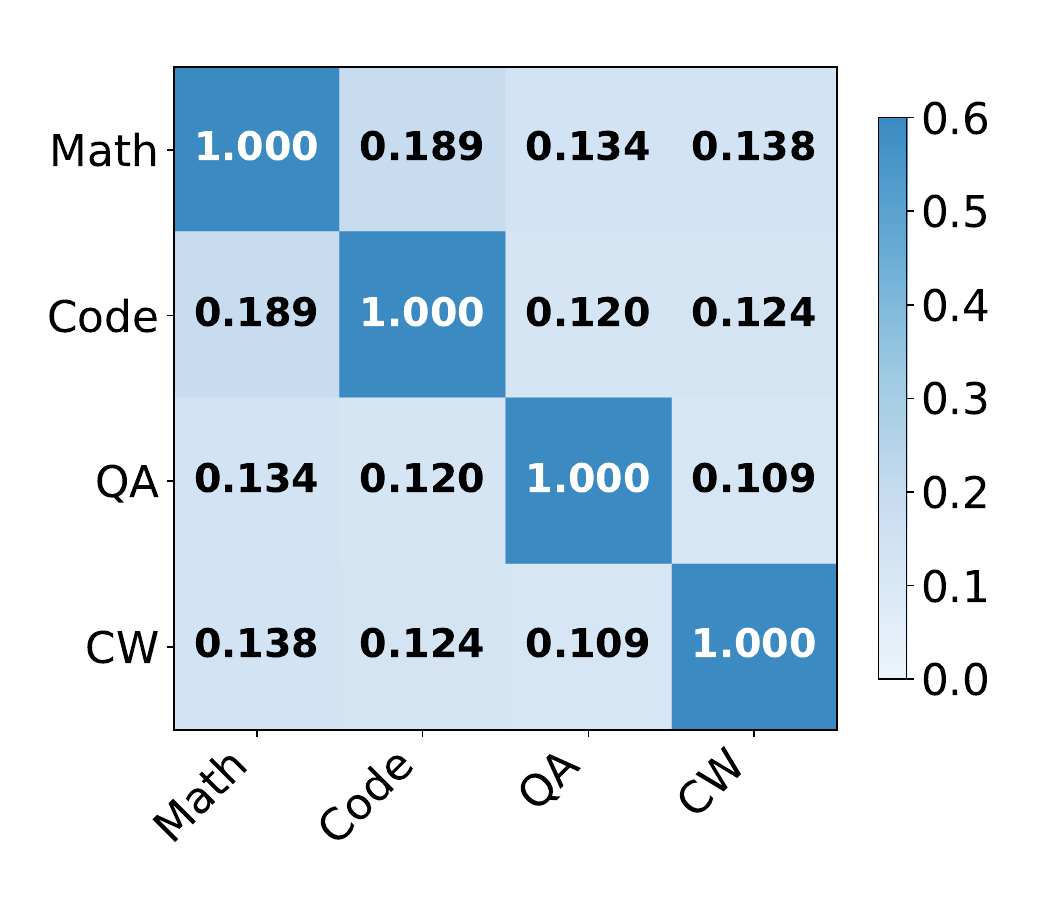}
    \vspace{-2em}
    \caption{Changed-neuron overlap.}
    \label{fig:jaccard_heatmap}
  \end{subfigure}
  \vspace{-1em}
  \begin{subfigure}[t]{\linewidth}
    \centering
    \includegraphics[width=\linewidth]{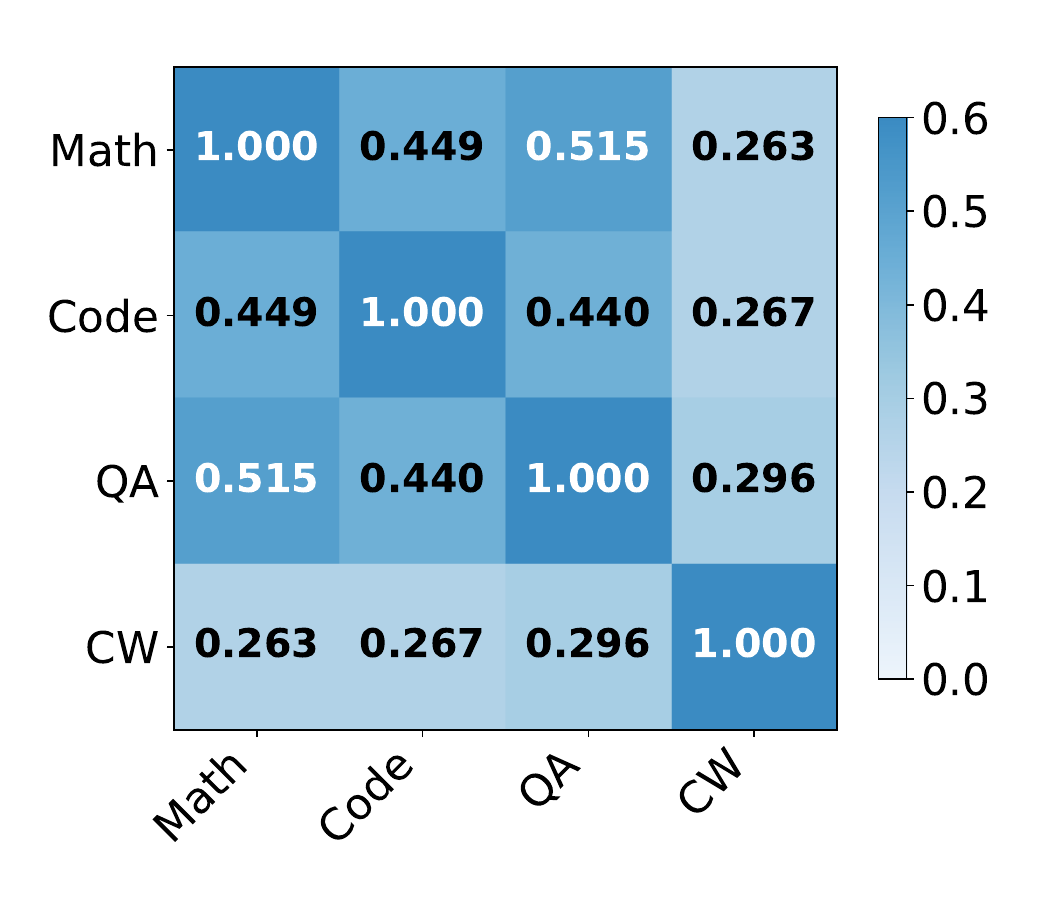}
    \vspace{-2em}
    \caption{Active-neuron overlap.}
    \label{fig:competition_heatmap}
  \end{subfigure}
  \vspace{-0.2em}
  \caption{Neuron-overlap rates under different settings.}
  \label{fig:overlap_comparison}
  \vspace{-4.5em}
\end{wrapfigure}

Gradients describe what training would do at an instant, not what was written into the weights. Following the distribution-shift view under which catastrophic forgetting is usually characterized, we compare each single-domain expert with the model it was trained from, measuring element-wise absolute and relative changes $\Delta \mW = \mW_{\mathrm{expert}} - \mW_{\mathrm{base}}$ and $r(\mW)=|\Delta \mW|/|\mW_{\mathrm{base}}|$. Across the four experts, 77\%--89\% of parameters have $|\Delta \mW|<10^{-7}$ and $r(\mW)<10^{-3}$ (Figure~\ref{fig:param_abs_per_task}), and even the parameters that do move move only slightly. Single-domain RL therefore acts as a mild perturbation of the base model rather than a global parameter rewrite. The relative-change distributions and the parameter analysis of sequential training are in Appendix~\ref{app:para}.

\subsection{Domains Edit Weakly Overlapping Neurons}
\label{subsec:edit_overlap}



Since domain RL updates are sparse, a natural explanation for strong interference is direct co-editing: different domains may concentrate their large updates on the same set of functional units. We test this explanation by lifting the analysis from parameters to MLP neurons and measuring the overlap among the most strongly changed neurons across domain experts. For each layer, we select the top 10\% most changed neurons for each domain expert and compute the pairwise Jaccard overlap between these sets. Full definitions are provided in Appendix~\ref{app:neuron_definition}.


Figure~\ref{fig:jaccard_heatmap} shows that top-changed neurons overlap only weakly across domains, with average Jaccard coefficients below 0.19 for all domain pairs. Domain RL therefore targets largely distinct local neuron subsets. This makes broad co-editing an incomplete description of the observed interference, although the small intersection may still matter. The next two subsections examine whether these edits act on functionally shared routes and whether their directions help characterize transfer and conflict.



\subsection{Sparse Edits Share Computation Routes}
\label{subsec:shared_routes}

Performance damage is itself evaluated at inference time, which makes it natural to shift the view from the parameters that were changed to the units that are used: low edit overlap does not imply functional independence, since two domains can write to almost disjoint weights and still read from the same units at inference. Using the same MLP-neuron definition as above, we rank neurons in each layer by their average activation magnitude on each domain, select the top 10\% most active neurons, and compute the pairwise Jaccard overlap of these active-neuron sets across domains. Full metric definitions are provided in Appendix~\ref{app:active_route_definition}.


Figure~\ref{fig:competition_heatmap} shows that active-neuron overlap among reasoning-oriented domains is far higher than changed-neuron overlap. Math-QA is the most coupled pair at 0.515, whereas CW overlaps with the other three domains only at 0.263--0.296. This ranking mirrors the training-time selective damage of Table~\ref{tab:task_level_main}: Math is damaged most by subsequent QA training, while CW is the most route-independent domain. We therefore localize cross-domain interference to sparse edits reaching another domain through the routes that both domains rely on at inference time.

\WFclear



\subsection{Edit Direction Governs Transfer vs.\ Conflict}
\label{subsec:directionality}

\begin{figure*}[t]
  \centering
  \includegraphics[width=0.95\linewidth]{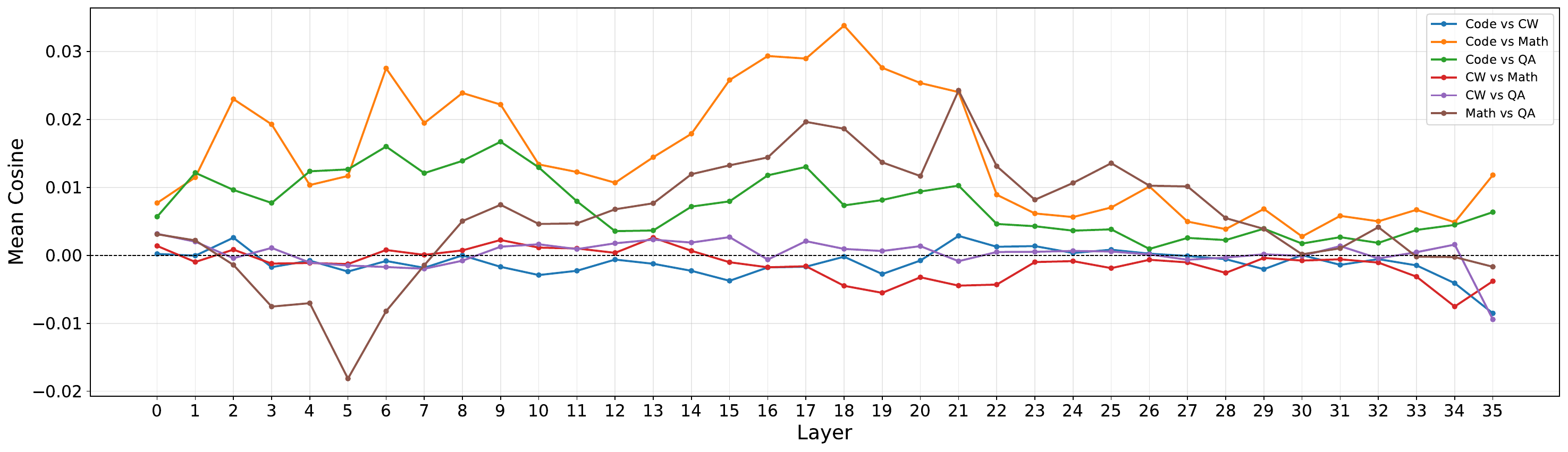}
  \caption{Layer-wise average directional cosine on shared top-changed neurons across domain pairs.}
  \label{fig:all_pairs_mean_cosine}
\end{figure*}



Section~\ref{subsec:shared_routes} identified shared active routes as the candidate site of interference, but overlap alone cannot distinguish transfer from damage: the missing ingredient is the \textit{direction} in which each domain moves a shared unit. We therefore restrict attention to the units that both domains actually edit. For each domain pair we intersect the two top-10\% changed-neuron sets and compute the cosine between the corresponding neuron-level update vectors (full definitions in Appendix~\ref{app:directional_alignment_metric}).

Figure~\ref{fig:all_pairs_mean_cosine} shows that domain pairs exhibit distinct directional patterns on these shared edited units. Code-Math is predominantly aligned, with the average cosine staying clearly positive across depth, consistent with the positive Code$\rightarrow$Math transfer in Table~\ref{tab:task_level_main}. Math-QA, by contrast, is not globally antagonistic but splits by depth: the average cosine is negative in layers L3--6 and positive in layers L14--21. Direction, rather than overlap, is what separates the synergistic pair from the damaging one, and it does so at a granularity that the module-level gradient view of Section~\ref{subsec:global_grad} cannot resolve.

\textbf{Three localization signals.} Together with the preceding subsections, this yields three measurable signals for selecting candidate repair coordinates: shared activation between the two domains (Section~\ref{subsec:shared_routes}), the magnitude of the later-domain edit (Section~\ref{subsec:sparse_update}), and directional conflict on the shared unit (this subsection). The same signals mark the coordinates that the direct intervention of Section~\ref{sec:proxy_conflict_subspace} later reverts; what they imply for how damage forms, and whether it can be undone, is the subject of the next section.
\section{A Local Perturbation Model of Interference and Repair}
\label{sec:theory}



The structural observations of Section~\ref{sec:ana} motivate a local perturbation model of interference. Shared active routes define candidate parameter subspaces, while edit magnitudes and directions motivate examining how displacement interacts with the earlier domain's loss geometry. We analyze fixed smooth objectives to separate first-order drift from curvature-sensitive loss changes and derive conditions under which a short refresh reduces the displacement on a shared subspace. This framework connects the structure of a later-domain update to the local sensitivity of an earlier-domain objective.

\subsection{Notation and Structural Assumptions}

Let $L_d(\vtheta)$ be a local objective for domain $d$, with gradient $\vg_d$ and Hessian $\mH_d$. After training on domain $A$, later training on $B$ induces displacement $\vdelta_B$ from the reference checkpoint $\vtheta_A^*$. We define interference by
\[
\Delta_{A\leftarrow B}=L_A(\vtheta_A^*+\vdelta_B)-L_A(\vtheta_A^*).
\]
The damage identities require local smoothness. Approximate stationarity bounds the first-order term, while the refresh result additionally assumes positive curvature on the relevant subspace, bounded coupling, and a region containing the iterates. These conditions are stated in Appendix~\ref{app:theory_assumptions}.

\paragraph{From shared routes to a parameter subspace.} Each unit $i$ in layer $\ell$ is associated with parameter coordinates $\mathcal C_i^{(\ell)}$ (gate and up rows and the down column for an MLP neuron; attention units analogously). For a set $R_{A,B}$ of shared active units edited by the later domain, define
\begin{equation}
S_{A,B}=\mathrm{span}\Bigl\{\ve_j:j\in\bigcup_{(i,\ell)\in R_{A,B}}\mathcal C_i^{(\ell)}\Bigr\},
\label{eq:subspace_def}
\end{equation}
and let $\mP_S$ denote its orthogonal projector. The empirical selector is a coordinate proxy for such routes. Its actual rollback operator, defined below, also allows partial reversion of shared attention weights. We first express the total loss change in terms of drift and curvature (Proposition~1), decompose its curvature contribution along the rollback operator (Proposition~2), and then analyze refresh on $S_{A,B}$ (Theorem~1).

\subsection{First-Order Drift and Curvature-Sensitive Damage}

Consider first the total damage $\Delta_{A\leftarrow B}$. The relevant curvature is averaged along the displacement:
\begin{equation}
\bar\mH_A=2\int_0^1(1-s)\mH_A(\vtheta_A^*+s\vdelta_B)\,ds,
\qquad
\bar\kappa_A(\vu)=\frac{\vu^\top\bar\mH_A\vu}{\|\vdelta_B\|_2^2}.
\label{eq:path_curvature}
\end{equation}

\paragraph{Proposition 1.}
\label{prop:second_order_damage}
\textit{Under local smoothness,}
\begin{equation}
\Delta_{A\leftarrow B}=\underbrace{\vg_A(\vtheta_A^*)^\top\vdelta_B}_{a}
+\tfrac12\|\vdelta_B\|_2^2\bar\kappa_A(\vdelta_B).
\label{eq:prop1_main}
\end{equation}
\textit{If $\|\vg_A(\vtheta_A^*)\|_2\le\varepsilon_A$, then $|a|\le\varepsilon_A\|\vdelta_B\|_2$.}

This identity retains both contributions. Curvature dominates when $|a|$ is small relative to the full curvature term; near-stationarity alone does not determine their relative sizes. Diagonalizing $\bar\mH_A$ gives a curvature contribution $\tfrac12\sum_k\bar\lambda_k c_k^2$, where $c_k$ is the displacement along eigenvector $k$, so a small update in sensitive directions can cause damage even when global gradient alignment is weak. The path average accounts for curvature variation along the displacement. The proof is in Appendix~\ref{app:proof_second_order}.

\subsection{Damage Decomposition Along a Sparse Intervention}
\label{sec:conflict_subspace}

Proposition~1 separates first-order drift from the total curvature contribution. To resolve the latter along a sparse intervention, we split the displacement into the component removed by a structured reversion and the component retained. Let $\mW$ be the diagonal rollback operator: its entries are one for fully reverted coordinates, $k/g$ for shared key and value rows selected by $k$ of their $g$ query heads, and zero otherwise. Write
\begin{equation}
\vv=\mW\vdelta_B,\qquad\vr=(\mI-\mW)\vdelta_B,\qquad\vdelta_B=\vv+\vr,
\label{eq:route_split}
\end{equation}
and define
\begin{equation}
Q_v=\vv^\top\bar\mH_A\vv,\qquad C=\vv^\top\bar\mH_A\vr,\qquad Q_r=\vr^\top\bar\mH_A\vr.
\label{eq:three_Q}
\end{equation}
The reverted and retained update components can share coordinates under partial rollback; no orthogonality is required.

\paragraph{Proposition 2.}
\label{prop:conflict_subspace}
\textit{Under local smoothness,}
\begin{equation}
\Delta_{A\leftarrow B}=a+\tfrac12Q_v+C+\tfrac12Q_r.
\label{eq:route_decomposition}
\end{equation}
\textit{Consequently, if $Q_v>0$,}
\begin{equation}
\left|\frac{\Delta_{A\leftarrow B}}{Q_v/2}-1\right|
\le\underbrace{\frac{2|a|}{Q_v}}_{\eta_1}
+\underbrace{\frac{2|C|+|Q_r|}{Q_v}}_{\eta_{\mathrm{loc}}}.
\label{eq:localization_certificate}
\end{equation}

Equation~\ref{eq:localization_certificate} bounds the discrepancy between the total loss change and the selected curvature contribution. The selected contribution approximates the total when the first-order term, the retained curvature, and the interaction term are small relative to $Q_v$. The decomposition thus identifies the contributions relevant to localization, rather than establishing it from sparsity alone. In particular, $C$ captures interaction with the retained update: the selected component cannot generally be assessed in isolation. This is an accounting of the original displacement, not a rollback-gain formula, since the remaining displacement $\vr$ follows a different path from the reference checkpoint. We evaluate the rollback's effect on task performance in Section~\ref{sec:proxy_conflict_subspace} and examine the selected direction's local sensitivity in Section~\ref{sec:curvature_validation}. The proof is in Appendix~\ref{app:proof_conflict_subspace}.

\subsection{A Conditional Explanation for Short Refresh}
\label{sec:refresh_contraction}

The preceding decomposition characterizes how the later-domain displacement changes the earlier objective. We now consider repair by resuming optimization on $A$ for a short time. Starting from $\vtheta_0=\vtheta_A^*+\vdelta_B$, we model refresh as local gradient updates $\vtheta_{t+1}=\vtheta_t-\alpha\vg_A(\vtheta_t)$. Positive curvature on $S_{A,B}$ and bounded coupling from its complement yield the following bound.

\paragraph{Theorem 1.}
\label{thm:refresh_contraction}
\textit{Under Assumptions~1--3 of Appendix~\ref{app:theory_assumptions}, with $0<\alpha\le1/\bar\beta_A$,}
\begin{equation}
\|\mP_S(\vtheta_t-\vtheta_A^*)\|_2
\le(1-\alpha\mu_A)^t\|\mP_S\vdelta_B\|_2
+\frac{\varepsilon_A+\xi_A R}{\mu_A}.
\label{eq:contraction}
\end{equation}

Here $\mu_A$ and $\bar\beta_A$ bound the projected average curvature, $\xi_A$ bounds cross-subspace coupling, and $R$ is an assumed radius containing the refresh trajectory. The bound combines a geometrically decaying term with an additive allowance for nonstationarity and coupling. This allowance controls the upper bound on the projected displacement; the actual displacement can be smaller. For a fixed admissible step size, larger $\mu_A$ gives faster decay, whereas stronger coupling increases the allowance. Thus, recovery depends on the subspace's loss geometry, not merely its size. The result gives a conditional explanation for rapid local recovery under gradient descent, not a quantitative guarantee for our adaptive-optimizer RL runs. Section~\ref{sec:exp_refresh} evaluates short refresh in our RL setting, and Appendix~\ref{app:proof_refresh} gives the proof.

With $G_A=\sup_{t<T}\|\vg_A(\vtheta_t)\|_2$, these idealized updates travel at most $\alpha TG_A$ from their starting point. Smoothness then controls changes to another domain.

\paragraph{Proposition 3.}
\label{prop:bounded_collateral}
\textit{If $L_C$ is $\beta_C$-smooth and $\|\vg_C(\vtheta_0)\|_2\le\varepsilon_C$, a $T$-step refresh satisfies}
\begin{equation}
L_C(\vtheta_T)-L_C(\vtheta_0)
\le\varepsilon_C\alpha TG_A+\tfrac12\beta_C\alpha^2T^2G_A^2.
\label{eq:collateral_bound}
\end{equation}

This bound permits small collateral changes when the displacement and the other domain's local sensitivity are small. Longer refresh tightens the contraction bound but increases this collateral bound, motivating a trade-off rather than prescribing an optimal duration; we choose the refresh length empirically (Appendix~\ref{app:refresh_dynamics}). Appendix~\ref{app:proof_other_domains} gives the proof, and Appendix~\ref{app:proof_alt_refresh} extends the first-order view to repeated alternating refresh over several domains. Propositions~1--2 characterize loss changes along a later-domain displacement, while Theorem~1 gives conditions for contracting its projection onto a shared subspace. We next evaluate refresh and rollback through task performance and probe the local sensitivity of the selected direction.

\begin{table*}[t]
\centering
\small
\caption{Performance of single-domain experts, sequential training, mixed-domain baselines}
\label{tab:task_level_main}
\resizebox{\textwidth}{!}{
\begin{tabular}{l|c|cccc|cccc|cccc}
\toprule
 Task
 & Base
 & $\mathrm{Code}_s$ & $\mathrm{Math}_s$ & $\mathrm{QA}_s$ & $\mathrm{CW}_s$
 & $\mathrm{Code}_o$ & $\mathrm{Math}_o$ & $\mathrm{QA}_o$ & $\mathrm{CW}_o$
 & CGPO & JT & Replay-all & Re-Math \\
\midrule
Math
& 43.19 & 59.63 & \textbf{66.84} & 55.31 & 39.78
& 59.63 & 66.49 & 59.90 & 57.66
& 61.93 & 64.80 & 63.54 & 66.04 \\
Code
& 29.57 & \textbf{52.67} & 34.65 & 32.07 & 28.15
& \textbf{52.67} & 50.69 & 50.99 & 50.47
& 50.05 & 48.61 & 51.09 & 51.05 \\
QA
& 60.64 & 60.89 & 60.76 & \textbf{63.31} & 60.50
& 60.89 & 60.52 & 62.34 & 62.34
& 62.48 & 62.11 & 62.56 & 62.49 \\
CW
& 82.44 & 82.40 & 81.38 & 81.76 & 86.24
& 82.40 & 81.44 & 81.79 & 86.52
& 86.73 & \textbf{86.97} & 86.08 & 85.96 \\
\midrule
AVG
& 53.96 & 63.90 & 60.91 & 58.11 & 53.67
& 63.90 & 64.79 & 63.76 & 64.25
& 65.30 & 65.62 & 65.82 & \textbf{66.39} \\
\bottomrule
\end{tabular}
}
\end{table*}

\section{Task-Level Validation and Direct Intervention}
\label{sec:exp}

We evaluate targeted repair through three experiments. Section~\ref{sec:exp_refresh} measures recovery of the damaged domain and changes to the other domains after a short refresh. Section~\ref{sec:proxy_conflict_subspace} evaluates training-free rollback on a sparse coordinate proxy for $S_{A,B}$. Section~\ref{sec:curvature_validation} compares the Math objective's directional curvature along the selected rollback direction with matched random controls.



\subsection{Recovery by Short Refresh}
\label{sec:exp_refresh}

\begin{wraptable}{r}{0.58\textwidth}
\vspace{-1.2em}
\footnotesize
\centering
\setlength{\tabcolsep}{3.5pt}
\caption{Selective rollback on a coordinate proxy ($\mathrm{QA}_o\!\to\!\mathrm{Math}_o$); Joint MLP+Attn adds attention coordinates. Recovery is the fraction of the Math accuracy drop recovered.}
\label{tab:proxy_conflict_revert}
\begin{tabular}{lcccc}
\toprule
Selector & Budget & Math Avg & Recovery & $\Delta$ QA Avg \\
\midrule
$\mathrm{Math}_o$ & -- & 66.49 & 100.0\% & $-1.81$ \\
$\mathrm{QA}_o$ & -- & 59.90 & 0.0\% & $0.00$ \\
Random & 1.3\% & 59.49 & -6.3\% & $+0.01$ \\
A$\times$M & 1.3\% & 60.53 & 9.6\% & $+0.04$ \\
A$\times$C & 1.3\% & 60.55 & 9.9\% & $-0.05$ \\
M$\times$C & 1.3\% & 61.11 & 18.3\% & $+0.04$ \\
A$\times$M$\times$C & 1.3\% & \textbf{61.25} & \textbf{20.4\%} & $-0.06$ \\
Joint MLP+Attn & 9.8\% & \textbf{63.68} & \textbf{57.4\%} & $-0.13$ \\
\bottomrule
\end{tabular}
\vspace{-2.0em}
\end{wraptable}

\textbf{Setup.} Table~\ref{tab:task_level_main} reports the single-domain experts, the sequential checkpoints, the mixed baselines ($\mathrm{JT}$, $\mathrm{CGPO}$, and a replay-all mixed refresh over all four domains from $\mathrm{CW}_o$), and $\mathrm{Re\text{-}Math}$, a short Math-only refresh from the same $\mathrm{CW}_o$ checkpoint (benchmark-level results in Appendix~\ref{app:full_results}).

\textbf{Selective interference and fast recovery.} Math reaches 66.49 after Code $\rightarrow$ Math, close to its single-domain expert (66.84), then drops to 57.66 over the later QA and CW stages. A short Math refresh from $\mathrm{CW}_o$ restores Math to 66.04, with Code at 51.05, QA at 62.49, and CW at 85.96, giving the highest average score of 66.39. The large Math recovery and comparatively limited changes to the other domains support selective repair (Appendices~\ref{app:refresh_dynamics} and~\ref{app:asym_directionality}).

\begin{wraptable}{r}{0.55\textwidth}
\vspace{-1.2em}
\footnotesize
\centering
\setlength{\tabcolsep}{3.5pt}
\caption{Normalized Math curvature density along the proxy $\vdelta_S$ vs.\ two equal-budget random directions at four QA training checkpoints (units $10^{-3}$).}
\label{tab:curvature_validation}
\begin{tabular}{lccccc}
\toprule
Ckpt & $\kappa(\hat S)$ & $\kappa(R_{\mathrm{l}})$ & $\kappa(R_{\mathrm{g}})$ & $\nicefrac{\hat S}{R_{\mathrm{l}}}$ & $\nicefrac{\hat S}{R_{\mathrm{g}}}$ \\
\midrule
$\mathrm{Math}_o$ & $+3.23$ & $+0.72$ & $+0.43$ & \textbf{4.51} & \textbf{7.44} \\
step 150 & $+17.2$ & $+13.4$ & $+7.77$ & 1.28 & 2.21 \\
$\mathrm{QA}_o$ & $-8.69$ & $+7.47$ & $+4.35$ & -- & -- \\
step 450 & $-4.28$ & $-1.00$ & $-0.73$ & 4.26 & 5.86 \\
\bottomrule
\end{tabular}
\vspace{-1.0em}
\end{wraptable}

\textbf{Comparison with baselines.} Re-Math achieves a higher Math score and macro-average than the evaluated joint and mixed-refresh baselines. Replay-all retains slightly higher scores on the other three domains, so the comparison is a task trade-off rather than uniform dominance. These results support targeted refresh for the damaged domain under our evaluated budgets, though the local theory establishes no general ordering between targeted and mixed training. The recovery pattern also holds when excluding the validation benchmark AIME25 (Appendix~\ref{app:full_results}).

\textbf{Generalization across model families.} The pattern replicates on a Phi-family model: under a Math$\rightarrow$QA$\rightarrow$Re-Math schedule, later QA training again degrades Math---by a smaller but directionally consistent margin---and a short refresh restores it with QA essentially unchanged (Appendix~\ref{app:cross_family}).


\subsection{Conflict Subspace Proxy}
\label{sec:proxy_conflict_subspace}

The refresh results demonstrate recoverability. We next test whether a sparse coordinate intervention can recover part of the QA-induced Math accuracy drop without further training. Using the structural signals of Section~\ref{subsec:directionality}, we construct a coordinate proxy and directly evaluate its rollback response.

Concretely, we score each MLP neuron by shared activation under Math and QA (A), the magnitude of the QA update (M), and directional conflict between the Math and QA task vectors (C). Defining $\vdelta_{Q\mid M}=\mathrm{QA}_o-\mathrm{Math}_o$, the intervention is $\vtheta_{\mathrm{rev}}=\mathrm{QA}_o-\mW\vdelta_{Q\mid M}$. For MLP-only selection, $\mW=\mP_{\hat S}$ is a binary coordinate projector; the joint MLP+Attn version also permits fractional reversion of shared KV weights.


Table~\ref{tab:proxy_conflict_revert} shows selective recovery. The $1.3\%$ A$\times$M$\times$C intervention recovers $20.4\%$ of the QA-induced Math accuracy drop with a QA Avg change of $-0.06$, while the matched-budget random intervention slightly reduces Math performance. At the same $1.3\%$ budget, removing magnitude or directional conflict roughly halves recovery (A$\times$C $9.9\%$, A$\times$M $9.6\%$), while M$\times$C remains close at $18.3\%$. Layer allocation also matters: activation-only ranking matches the full selector only when inheriting its layer budgets (Appendix~\ref{app:proxy_conflict_appendix}). The joint intervention at $9.8\%$ recovers $57.4\%$ with a QA change of $-0.13$. The two budgets use different selectors, so the comparison does not establish a monotone relation between budget and recovery.

\subsection{Directional Curvature of the Conflict Proxy}
\label{sec:curvature_validation}

We examine the Math objective's directional curvature along the selected rollback direction and matched random directions. These checkpoint-level probes characterize the local sensitivity of the repair direction.

Fixing the actual rollback direction $\vdelta_S=\mW\vdelta_{Q\mid M}$ for the $9.8\%$ joint selection, we measure $\kappa(\hat S;\vtheta)=\vdelta_S^\top\mH_{\mathrm{Math}}(\vtheta)\vdelta_S/\|\vdelta_S\|_2^2$ by centered finite differences of Math GRPO gradients on a fixed diagnostic batch. The control $R_{\mathrm{l}}$ matches per-layer, per-module unit counts, while $R_{\mathrm{g}}$ matches the total MLP and attention counts (Appendix~\ref{app:curvature_details}). Their ratios compare the selected direction with these controls at each checkpoint.

\begin{wrapfigure}{r}{0.45\textwidth}
  \vspace{-1.2em}
  \centering
  \includegraphics[width=\linewidth]{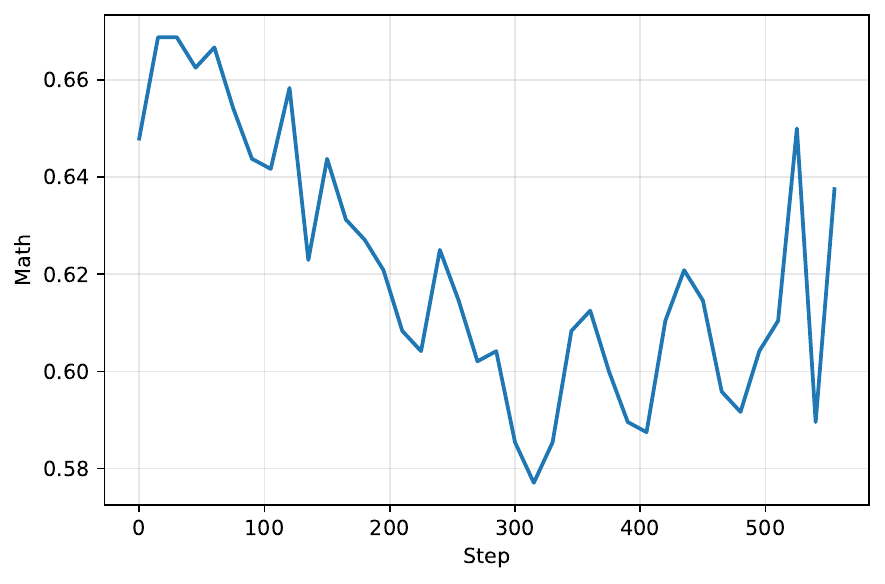}
  \caption{Math validation accuracy over QA training. Directional curvature at four checkpoints is reported in Table~\ref{tab:curvature_validation}.}
  \label{fig:curvature_vs_math}
  \vspace{-2.0em}
\end{wrapfigure}

\textbf{Higher curvature along the selected direction.} At $\mathrm{Math}_o$, curvature along $\vdelta_S$ is $4.5\times$ that of $R_{\mathrm{l}}$ and $7.4\times$ that of $R_{\mathrm{g}}$ (Table~\ref{tab:curvature_validation}). Reducing the probe step from $10^{-3}$ to $3\times10^{-4}$ changes the absolute estimate, while the selected direction retains approximately $4\times$ and $5\times$ the curvature of the respective controls (Appendix~\ref{app:curvature_details}). The comparison with the layer-matched control supports sensitivity within layers as well as across them. Combined with the rollback results, these probes support the selected direction as a useful repair target.

\textbf{Curvature varies along the QA trajectory.} The selected direction has positive Math curvature at $\mathrm{Math}_o$ and step 150, and negative curvature at $\mathrm{QA}_o$ and step 450. Table~\ref{tab:curvature_validation} and Figure~\ref{fig:curvature_vs_math} describe this evolving local geometry alongside the Math validation trajectory. These probes characterize curvature at individual training checkpoints. The loss decomposition in Proposition~2 uses the average Hessian along the parameter segment joining $\mathrm{Math}_o$ and $\mathrm{QA}_o$, whose value also depends on curvature between the endpoints.

\section{Conclusion}
\label{sec:conclusion}

Our results support a local sensitivity account of cross-domain interference in multi-domain RL. Sparse edits, shared active routes, and directional alignment reveal structure that full-model gradient comparisons alone do not capture. The local perturbation analysis provides an exact loss decomposition and a conditional explanation for short refresh. Empirically, targeted refresh recovers the damaged Math domain with limited changes elsewhere, and sparse training-free rollback also restores a substantial part of the accuracy drop. Curvature probes provide complementary evidence that the selected directions are locally sensitive. Together, these findings motivate targeted intervention as a practical way to mitigate cross-domain degradation.

Our experiments use single training runs, so small score differences should be interpreted cautiously. The coordinate proxy is one practical intervention and does not identify a minimal conflict subspace. The theory is conditional on local smoothness, curvature, and coupling assumptions; its loss-level bounds are not quantitative predictions of benchmark accuracy. Theorem~1 and Proposition~3 analyze gradient descent on fixed local objectives, whereas our RL runs use an adaptive optimizer and a moving policy-gradient surrogate. Our finite-difference probes characterize checkpoint-level directional sensitivity, and their absolute magnitudes depend on the perturbation step. Measuring the path-averaged loss decomposition, extending the refresh analysis to adaptive policy optimization, and automatically selecting repair directions and budgets are directions for future work.


%

\section*{Ethics Statement}

This work is a diagnostic study of cross-domain interference in multi-domain RL post-training. All experiments use a publicly available model (Qwen3-4B-Thinking-2507) and public datasets (OpenR1-Math, KlearReasoner-CodeSub-15K, SuperGPQA, and the Creative-Writing series), and involve no human subjects, private data, or personally identifiable information. Our analysis and the proposed targeted refresh introduce no new generative capabilities beyond standard domain RL; rather, they aim to make multi-domain post-training more stable and compute-efficient, which we regard as a positive broader impact. The standard risks of large language models---hallucination, and biased or unsafe generations---apply to the underlying model and are neither amplified nor specifically addressed here. We foresee no direct avenue for misuse of our findings.

\section*{Reproducibility Statement}

We provide the setup for the reported experiments: data construction and evaluation benchmarks (Appendix~\ref{app:data_const}), training hyperparameters (Appendix~\ref{app:train_hyper}), and per-domain rewards (Appendix~\ref{app:reward_func}). Definitions of changed neurons, active routes, and directional alignment are given in Appendices~\ref{app:neuron_definition}, \ref{app:active_route_definition}, and~\ref{app:directional_alignment_metric}. The local theory's assumptions and proofs are stated in Appendix~\ref{app:det_proof}. The rollback selector, finite-difference curvature estimator, and matched random controls are described in Appendices~\ref{app:proxy_conflict_appendix} and~\ref{app:curvature_details}. We will release the training and analysis code together with the domain configurations.

\section*{AI Use Statement}

In accordance with the ICLR 2027 AI Policy for Authors, we disclose the following uses of generative AI. Qwen3-235B-A22B-Instruct-2507 was used to generate replacement reference responses for 2,560 creative-writing training examples (Appendix~\ref{app:data_const}) and as an automated judge for the creative-writing reward during training and evaluation (Appendix~\ref{app:reward_func}). Large language model assistants were additionally used for writing and coding assistance. The authors retain responsibility for the research design, experiments, verification of all claims and proofs, and the final content of this paper, including all AI-assisted text, code, data artifacts, and evaluations.

\bibliographystyle{iclr2027_conference}
\bibliography{references}


\clearpage

\appendix

\begin{table*}[!t]
\centering
\footnotesize
\caption{Training hyperparameters.}
\label{tab:selected_hyperparams}
\resizebox{\textwidth}{!}{%
\begin{tabular}{cccccc}
\toprule
\textbf{train batch size} & \textbf{ppo mini batch size} & \textbf{max prompt length} & \textbf{max response length} & \textbf{adv estimator} & \textbf{filter overlong prompts} \\
\midrule
256 & 256 & 2048 & 16384 & grpo & True \\
\midrule
\textbf{lr} & \textbf{use dynamic bsz} & \textbf{use kl loss} & \textbf{kl loss coef} & \textbf{kl loss type} & \textbf{clip ratio} \\
\midrule
1e-6 & True & False & 0.0 & low\_var\_kl & 0.2 \\
\midrule
\textbf{entropy coeff} & \textbf{rollout n} & \textbf{rollout temperature} & \textbf{rollout top k} & \textbf{rollout top p} & \textbf{val top k} \\
\midrule
0 & 8 & 1.0 & -1 & 1.0 & -1 \\
\midrule
\textbf{val top p} & \textbf{val temperature} & \textbf{val n} & \textbf{val do sample} & \textbf{use kl in reward} & \\
\midrule
0.9 & 0.7 & 8 & True & False & \\
\bottomrule
\end{tabular}
}
\end{table*}

\section{Implementation Details}

\subsection{Data Construction Details}
\label{app:data_const}

We ensure that the prompt length of all training data is within 2,048. Each domain uses 5,120 training examples. Validation sets are constructed from non-training examples in the corresponding data sources. For Math, we use AIME25~\citep{aops_aime} as the validation set, which contains 30 problems. For Code, QA, and CW, each validation set contains 50 examples.

For CW, we sample 5,120 examples from the crownelius/Creative-Writing series.\footnote{\url{https://huggingface.co/Crownelius}} The data come from four subsets: Sonnet4.6-800x,\footnote{\url{https://huggingface.co/datasets/Crownelius/Creative-Writing-Sonnet4.6-800x}} Gemini3Pro-2700x,\footnote{\url{https://huggingface.co/datasets/Crownelius/Creative-Writing-Gemini3Pro-2700x}} Reasoning-KimiK2.5-600x,\footnote{\url{https://huggingface.co/datasets/Crownelius/Creative-Writing-Reasoning-KimiK2.5-600x}} and Qwen3.5Plus-2000x.\footnote{\url{https://huggingface.co/datasets/Crownelius/Creative-Writing-Qwen3.5Plus-2000x}} For 2,560 of these examples, we resample the reference responses using Qwen3-235B-A22B-Instruct-2507 and use the newly sampled responses as references.

For evaluation, LiveCodeBench-v6 contains 175 problems spanning January 2025 to April 2025. For QA evaluation, after excluding the training and validation data from SuperGPQA, we sample 10\% of the remaining data, resulting in 2,141 test examples. Similarly, we sample 10\% of MMLU-Pro, resulting in 1,203 test examples.


\subsection{Training Hyperparameters}
\label{app:train_hyper}

All experiments use the same GRPO training configuration. The full hyperparameters are summarized in Table~\ref{tab:selected_hyperparams}.

At each training stage, we train the model until convergence and select the checkpoint with the optimal validation performance in the current training domain. Specifically, single-domain training adopts the best-performing checkpoint within the single domain, while multi-domain mixed training selects the checkpoint with the optimal average performance across all domains.


\subsection{Reward Functions}
\label{app:reward_func}

Math and QA use answer-correctness rewards. The model is required to put the final answer in boxed. We use math\_verify to parse the answer and match it against the reference answer. The reward is binary: 1 for a correct answer and 0 otherwise.

Code uses execution-based rewards, computed from the fraction of test cases passed.

CW uses an LLM-as-a-judge preference reward. For crownelius/Creative-Writing series, given a prompt, a reference response, and a model response, the judge compares their quality. If the model response is better, the reward is 1; if the two responses are comparable, the reward is 0.5; if the reference response is better, the reward is 0. For each query of WritingBench, the model generates one response, which is scored along multiple WritingBench evaluation dimensions. We use Qwen3-235B-A22B-Instruct-2507~\citep{DBLP:journals/corr/abs-2505-09388} as the judge for training and evaluation.

\begin{figure*}[t]
  \centering
  \begin{subfigure}[t]{0.48\textwidth}
    \centering
    \includegraphics[width=\linewidth]{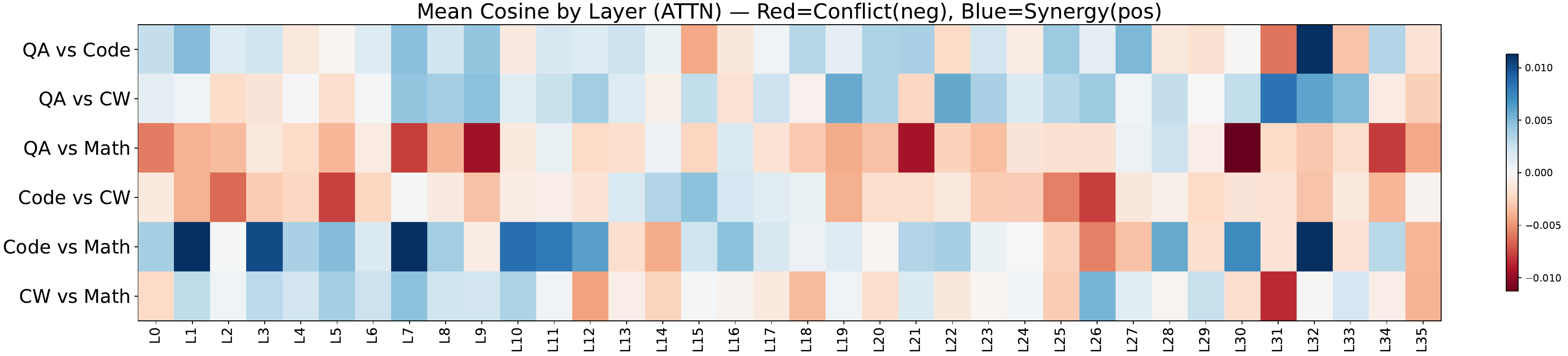}
    \caption{Attention-module heatmap of pairwise gradient cosine.}
    \label{fig:cosine_heatmap_attn}
  \end{subfigure}\hfill
  \begin{subfigure}[t]{0.48\textwidth}
    \centering
    \includegraphics[width=\linewidth]{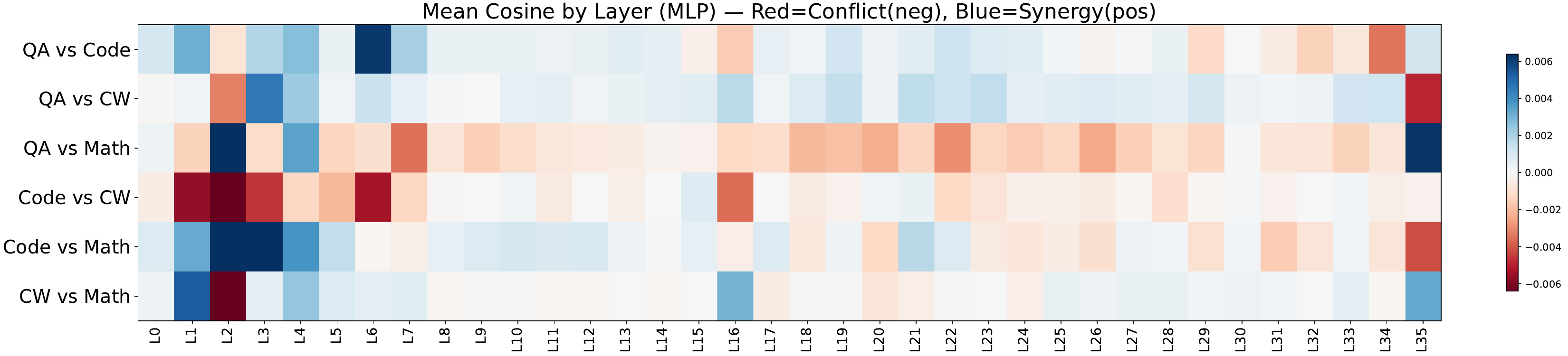}
    \caption{MLP-module heatmap of pairwise gradient cosine.}
    \label{fig:cosine_heatmap_mlp}
  \end{subfigure}
  \caption{Layer-wise heatmaps of pairwise gradient cosine in attention and MLP modules.}
  \label{fig:grad_conflict_heatmaps}
\end{figure*}

\begin{figure*}[t]
  \centering
  \begin{subfigure}[t]{0.48\textwidth}
    \centering
    \includegraphics[width=\linewidth]{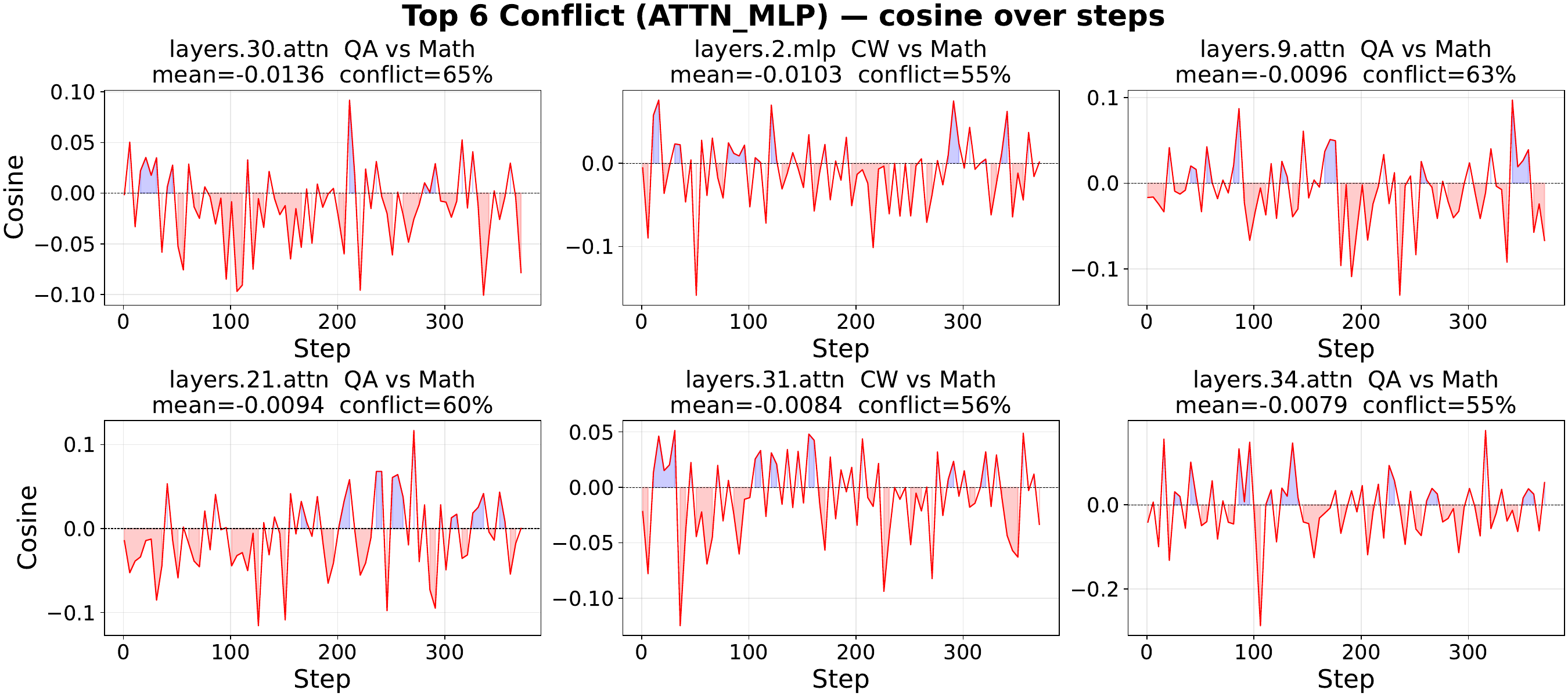}
    \caption{Top six attention and MLP modules with the strongest conflict.}
    \label{fig:grad_conflict_top6}
  \end{subfigure}\hfill
  \begin{subfigure}[t]{0.48\textwidth}
    \centering
    \includegraphics[width=\linewidth]{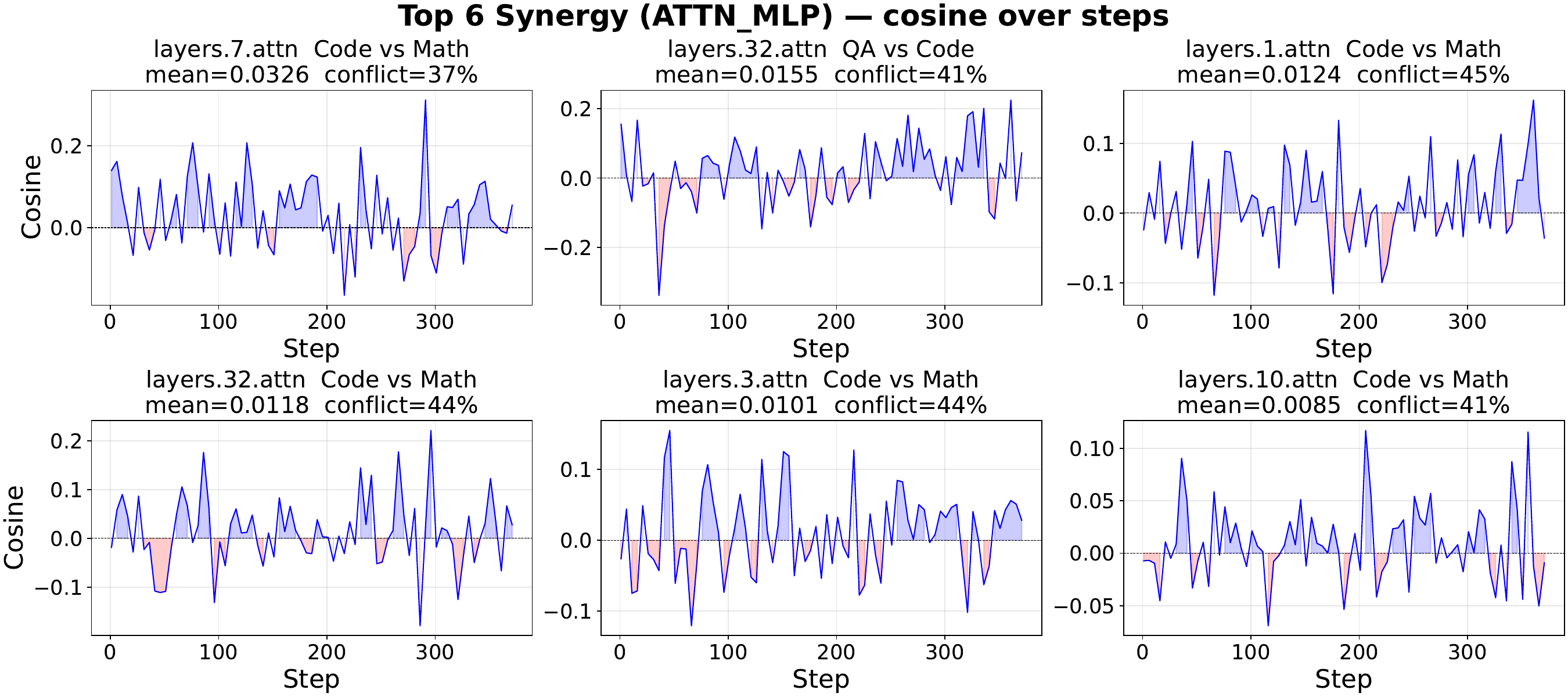}
    \caption{Top six attention and MLP modules with the strongest synergy.}
    \label{fig:grad_synergy_top6}
  \end{subfigure}
  \caption{Module-level conflict and synergy trends across the most prominent attention and MLP modules.}
  \label{fig:grad_conflict_trends}
\end{figure*}

\section{Additional Structural Analysis}

\subsection{Extended Module-Level Gradient Conflict and Synergy}
\label{app:grad_conf}

Figure~\ref{fig:grad_conflict}(b,c) in the main text tracks the single most conflicting and most synergistic module for the Math-QA pair over joint training. Figure~\ref{fig:grad_conflict_trends} extends this to the six attention and MLP modules with the most prominent conflict and synergy, respectively, and Figure~\ref{fig:grad_conflict_heatmaps} presents the layer-wise heatmaps of pairwise gradient cosine in attention and MLP modules. These results show that conflicts are not uniformly distributed across all parameters but are concentrated in specific regions of the network. In particular, local conflicts are most significant for the Math-QA domain pair. In addition, certain local conflicts also exist between Code-CW. Nevertheless, despite relatively clear module interactions, the corresponding cosine values remain small and are largely orthogonal, which is why Section~\ref{sec:ana} treats this module-level decomposition only as a hint rather than as a localization of the interference.


\subsection{Sequential Parameter Changes Remain Sparse}
\label{app:para}

\begin{figure}[h]
  \centering
  \includegraphics[width=0.48\textwidth]{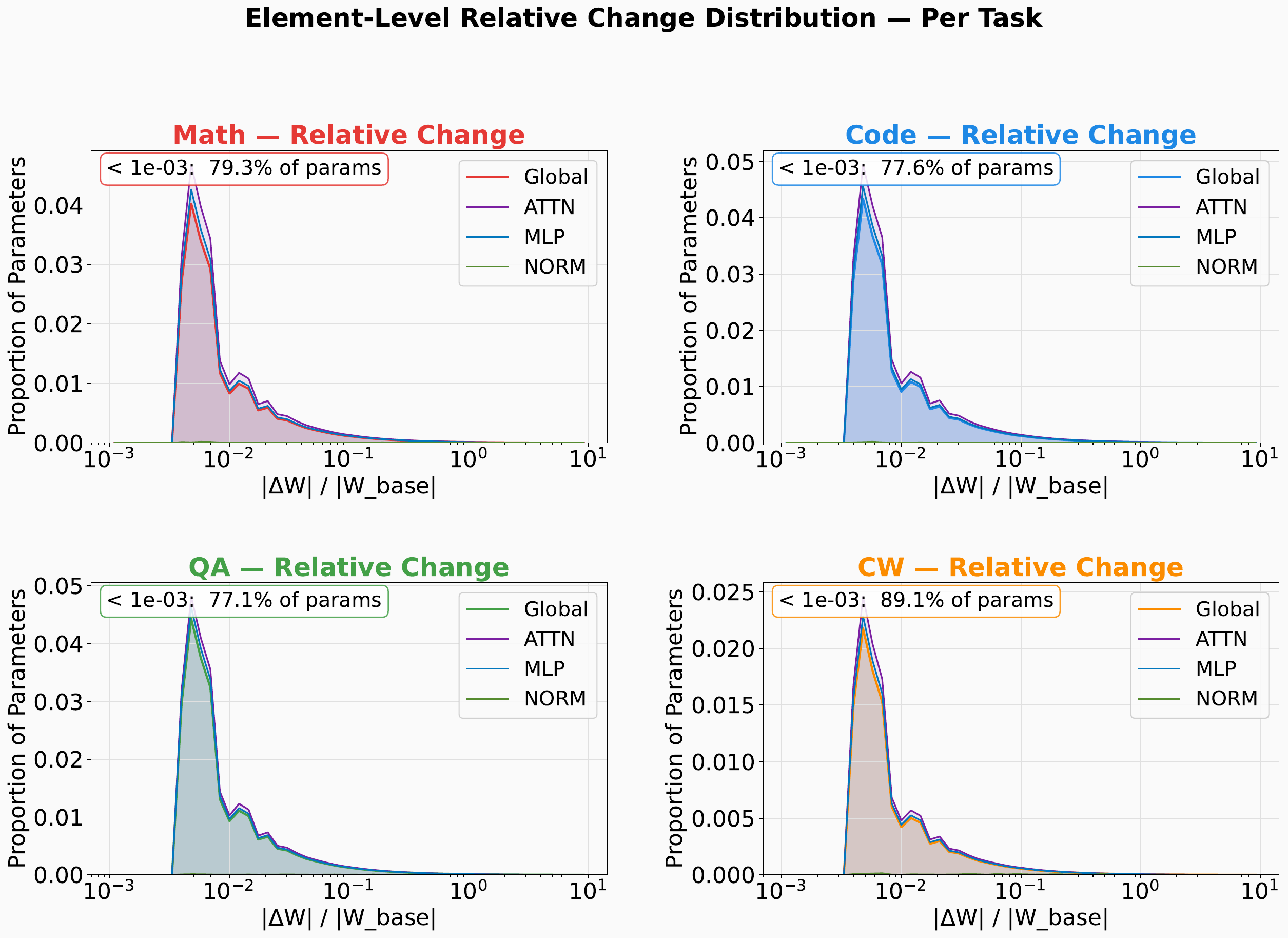}
  \caption{Relative parameter changes $r(\mW)$ of the four single-domain experts relative to the base model, the companion of Figure~\ref{fig:param_abs_per_task} omitted from Section~\ref{subsec:sparse_update}.}
  \label{fig:param_rel_per_task}
\end{figure}

Figure~\ref{fig:param_rel_per_task} gives the relative-change distributions that accompany the absolute changes reported in Section~\ref{subsec:sparse_update}: the two views agree, with the bulk of the mass below $r(\mW)=10^{-3}$ for every expert, so neither the absolute nor the normalized scale reveals a global rewrite.

In addition to single-domain experts, we also measure parameter changes along the sequential training chain: Code $\rightarrow$ Math $\rightarrow$ QA $\rightarrow$ CW. We analyze both cumulative changes relative to the base model and incremental changes relative to the previous checkpoint.

\begin{figure*}[t]
  \centering
  \begin{subfigure}[t]{0.48\textwidth}
    \centering
    \includegraphics[width=\linewidth]{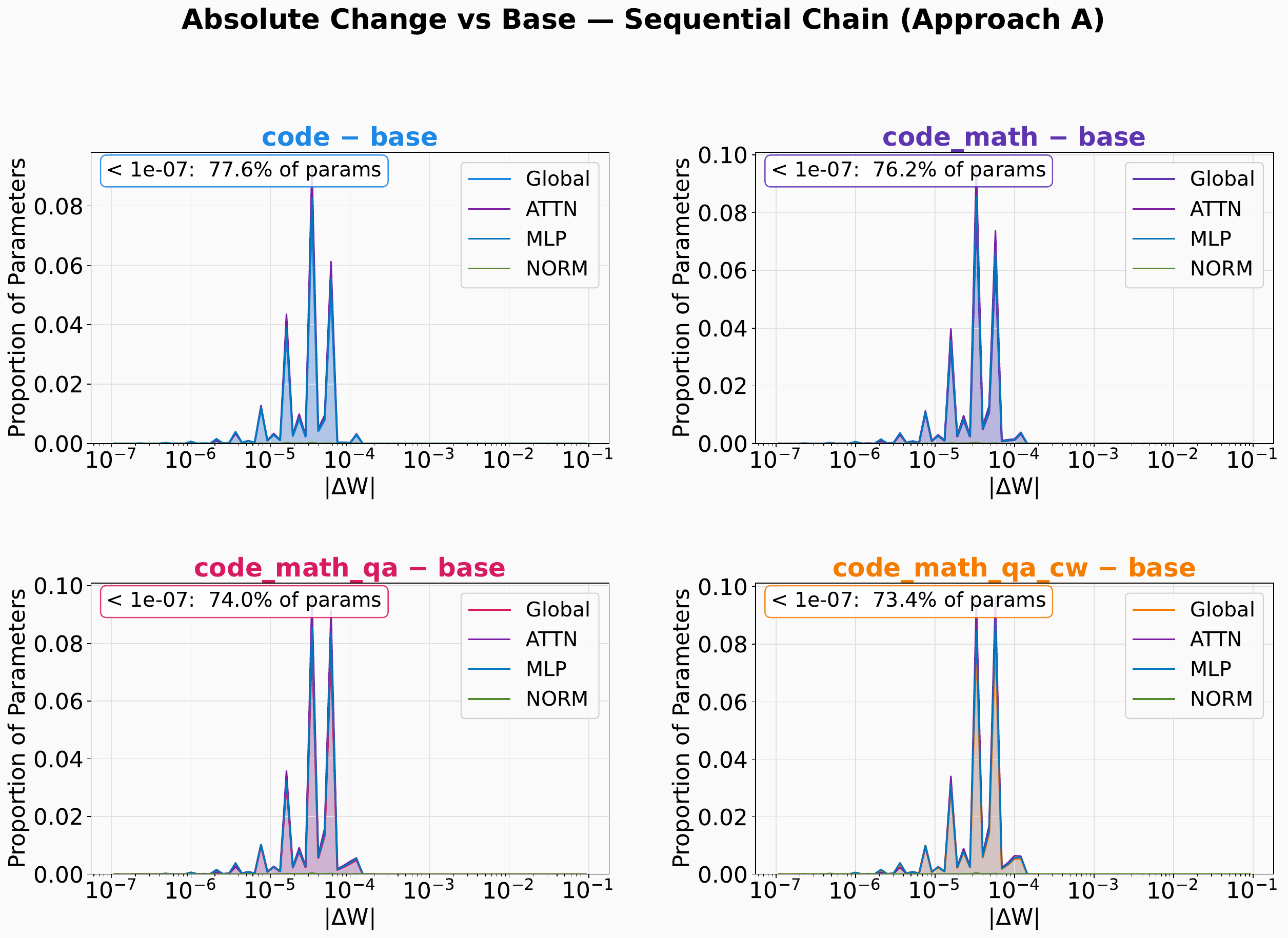}
    \caption{Absolute cumulative parameter changes relative to the base model.}
    \label{fig:param_abs_vs_base}
  \end{subfigure}\hfill
  \begin{subfigure}[t]{0.48\textwidth}
    \centering
    \includegraphics[width=\linewidth]{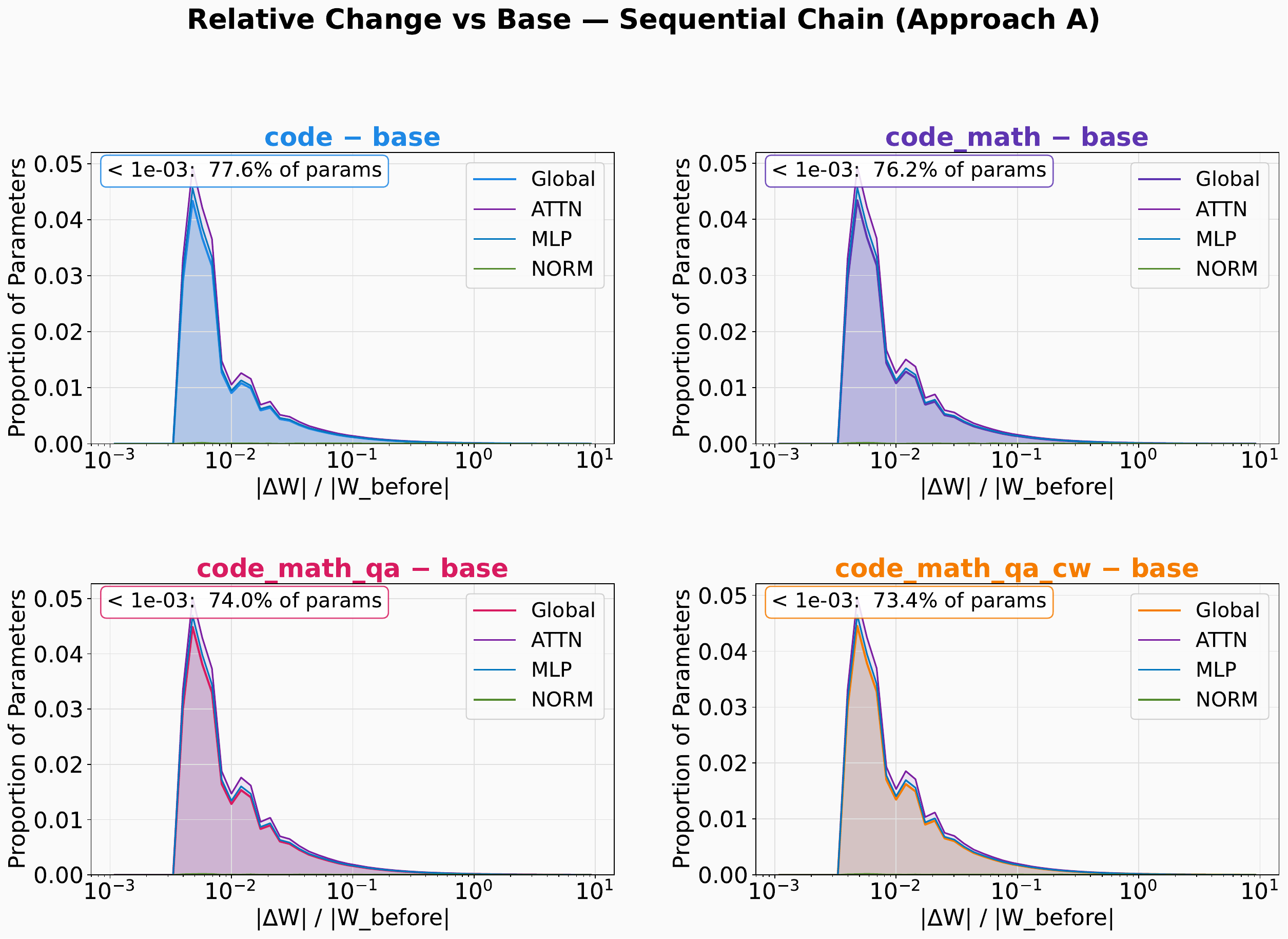}
    \caption{Relative cumulative parameter changes relative to the base model.}
    \label{fig:param_rel_vs_base}
  \end{subfigure}
  \caption{Cumulative parameter-change distributions along the sequential domain RL chain relative to the base model.}
  \label{fig:param_seq_vs_base}
\end{figure*}

As shown in Figure~\ref{fig:param_seq_vs_base}, cumulative deviation from the base model gradually increases as more domains are added. The fraction of parameters with $|\Delta W| < 10^{-7}$ and $r(W) < 10^{-3}$ decreases from 77.6\% after Code training to 73.4\% after Code-Math-QA-CW training.

\begin{figure*}[t]
  \centering
  \begin{subfigure}[t]{0.48\textwidth}
    \centering
    \includegraphics[width=\linewidth]{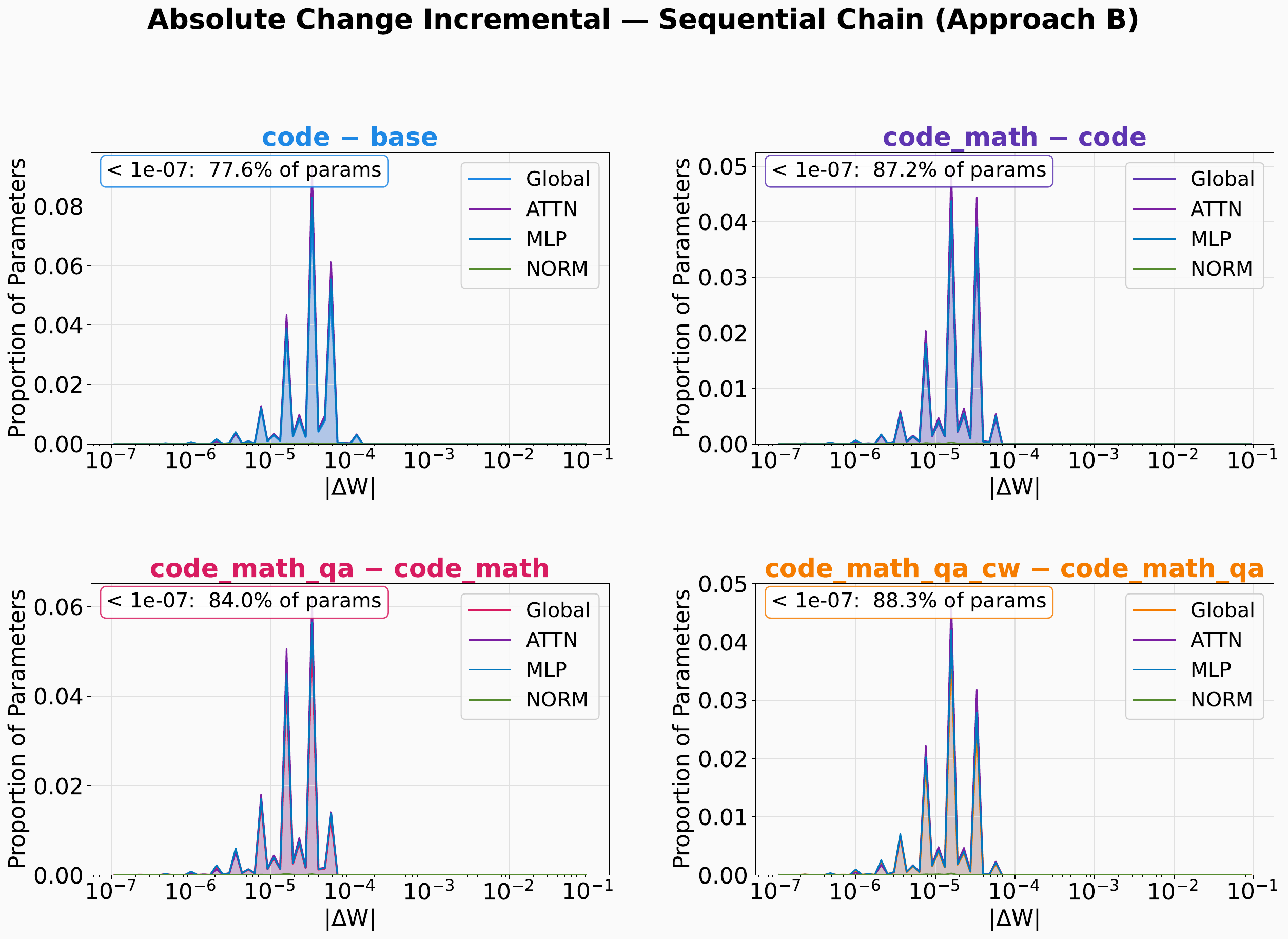}
    \caption{Absolute incremental parameter changes relative to the previous checkpoint.}
    \label{fig:param_abs_incremental}
  \end{subfigure}\hfill
  \begin{subfigure}[t]{0.48\textwidth}
    \centering
    \includegraphics[width=\linewidth]{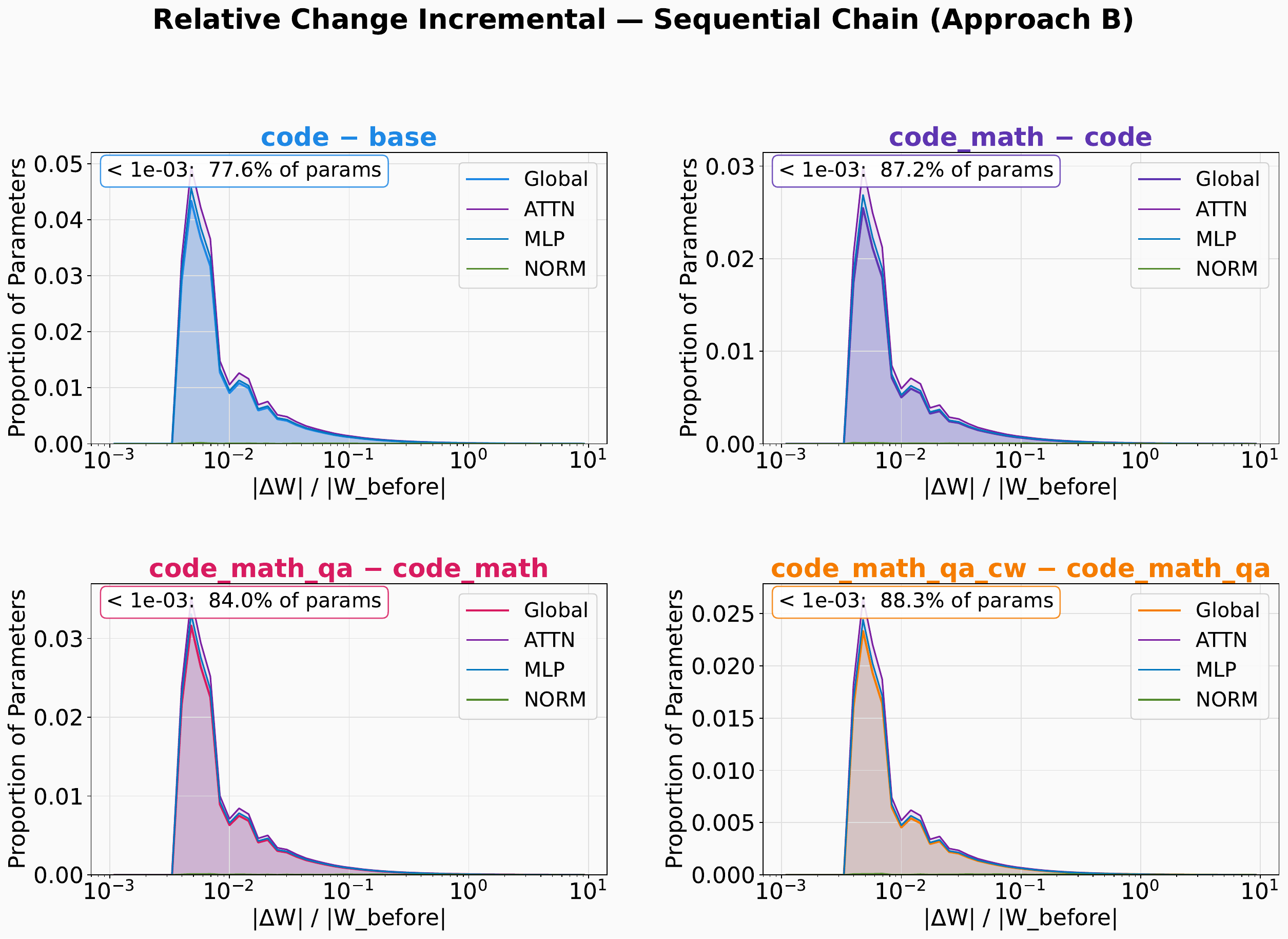}
    \caption{Relative incremental parameter changes relative to the previous checkpoint.}
    \label{fig:param_rel_incremental}
  \end{subfigure}
  \caption{Incremental parameter-change distributions at each stage of the sequential domain RL chain.}
  \label{fig:param_seq_incremental}
\end{figure*}

However, as shown in Figure~\ref{fig:param_seq_incremental}, each individual stage remains sparse. When comparing each checkpoint with the immediately preceding one, later domain stages modify only a small fraction of parameters: 87.2\%, 84.0\%, and 88.3\% of parameters remain below the $10^{-7}$ threshold for the Math, QA, and CW stages, respectively.

These results show that sequential domain RL accumulates parameter shifts over domains, but each stage still acts as a sparse incremental update rather than a global parameter rewrite.



\subsection{Neuron-Level Edit Definition}
\label{app:neuron_definition}

For each MLP layer $\ell$, we define the $i$-th intermediate channel as one neuron. Its associated parameters consist of the corresponding row of the gate projection, the corresponding row of the up projection, and the corresponding column of the down projection:
\begin{equation}
\vg_i^{(\ell)} = \mW_{\mathrm{gate}}^{(\ell)}[i,:], \quad \vu_i^{(\ell)} = \mW_{\mathrm{up}}^{(\ell)}[i,:], \quad \vd_i^{(\ell)} = \mW_{\mathrm{down}}^{(\ell)}[:,i].
\end{equation}

For domain $d$, let $\Delta \vg_{d,i}^{(\ell)}$, $\Delta \vu_{d,i}^{(\ell)}$, and $\Delta \vd_{d,i}^{(\ell)}$ denote the corresponding parameter changes between the domain expert and the base model. We measure the edit magnitude of neuron $i$ in layer $\ell$ by aggregating the squared parameter changes over these three components:
\begin{equation}
s_{d,i}^{(\ell)} = \left\| \Delta \vg_{d,i}^{(\ell)} \right\|_2^2 + \left\| \Delta \vu_{d,i}^{(\ell)} \right\|_2^2 + \left\| \Delta \vd_{d,i}^{(\ell)} \right\|_2^2.
\end{equation}
For each layer and each domain, we select the top 10\% neurons with the largest edit magnitude:
\begin{equation}
\mathcal{N}_d^{(\ell)} = \operatorname{Top}_{10\%} \left\{s_{d,i}^{(\ell)}\right\}_{i}.
\end{equation}
The layer-wise overlap between two domains $A$ and $B$ is measured by the Jaccard coefficient:
\begin{equation}
J^{(\ell)}(A,B) = \frac{ |\mathcal{N}_A^{(\ell)} \cap \mathcal{N}_B^{(\ell)}| }{ |\mathcal{N}_A^{(\ell)} \cup \mathcal{N}_B^{(\ell)}| }.
\end{equation}
Finally, we report the average overlap across the $L$ MLP layers:
\begin{equation}
J(A,B) = \frac{1}{L} \sum_{\ell=1}^{L} J^{(\ell)}(A,B).
\end{equation}

\subsection{Active-Route Overlap Metric}
\label{app:active_route_definition}

Using the same MLP-neuron definition as in Appendix~\ref{app:neuron_definition}, we measure whether two domains activate overlapping active routes during inference. For domain $d$ and layer $\ell$, let $h_{d,i,t}^{(\ell)}(x)$ denote the activation of neuron $i$ at token position $t$ for sample $x \in \mathcal{D}_d$, and let $T_x$ be the number of tokens in $x$. We define the sample-level average activation magnitude as
\begin{equation}
\bar{a}_{d,i}^{(\ell)}(x) = \frac{1}{T_x} \sum_{t=1}^{T_x} \left|h_{d,i,t}^{(\ell)}(x)\right|.
\end{equation}
The dataset-level activation score is
\begin{equation}
a_{d,i}^{(\ell)} = \frac{1}{|\mathcal{D}_d|} \sum_{x\in \mathcal{D}_d} \bar{a}_{d,i}^{(\ell)}(x).
\end{equation}
For each layer and each domain, we select the top 10\% neurons with the largest dataset-level activation score:
\begin{equation}
\mathcal{A}_d^{(\ell)} = \operatorname{Top}_{10\%} \left\{a_{d,i}^{(\ell)}\right\}_{i}.
\end{equation}
The layer-wise active-route overlap between two domains $A$ and $B$ is measured by the Jaccard coefficient:
\begin{equation}
J_{\mathrm{act}}^{(\ell)}(A,B) = \frac{ |\mathcal{A}_{A}^{(\ell)} \cap \mathcal{A}_{B}^{(\ell)}| }{ |\mathcal{A}_{A}^{(\ell)} \cup \mathcal{A}_{B}^{(\ell)}| }.
\end{equation}
Finally, we report the average active-route overlap across the $L$ MLP layers:
\begin{equation}
J_{\mathrm{act}}(A,B) = \frac{1}{L} \sum_{\ell=1}^{L} J_{\mathrm{act}}^{(\ell)}(A,B).
\end{equation}

\subsection{Directional Alignment on Shared Edited Neurons}
\label{app:directional_alignment_metric}

For each domain $d$, layer $\ell$, and MLP neuron $i$, we define the neuron-level update vector by concatenating the parameter changes associated with its gate, up, and down projections:
\begin{equation}
\Delta \vv_{d,i}^{(\ell)} = \left[ \Delta \vg_{d,i}^{(\ell)}; \Delta \vu_{d,i}^{(\ell)}; \Delta \vd_{d,i}^{(\ell)} \right].
\end{equation}
For two domains $A$ and $B$, we define the shared edited neuron set in layer $\ell$ as
\begin{equation}
\mathcal{Q}_{A,B}^{(\ell)} = \mathcal{N}_{A}^{(\ell)} \cap \mathcal{N}_{B}^{(\ell)},
\end{equation}
where $\mathcal{N}_{d}^{(\ell)}$ is the Top-10\% changed-neuron set defined in Appendix~\ref{app:neuron_definition}. For each shared edited neuron, we compute the cosine similarity between the two domain update vectors:
\begin{equation}
c_{A,B,i}^{(\ell)} = \frac{ \langle \Delta \vv_{A,i}^{(\ell)}, \Delta \vv_{B,i}^{(\ell)} \rangle }{ \|\Delta \vv_{A,i}^{(\ell)}\|_2 \|\Delta \vv_{B,i}^{(\ell)}\|_2 }.
\end{equation}
The layer-wise average directional alignment on shared edited neurons is
\begin{equation}
C^{(\ell)}(A,B) = \frac{1}{|\mathcal{Q}_{A,B}^{(\ell)}|} \sum_{i\in\mathcal{Q}_{A,B}^{(\ell)}} c_{A,B,i}^{(\ell)}.
\end{equation}
Positive values indicate that two domains edit their shared neurons in aligned directions on average, while negative values indicate conflicting directions.

\paragraph{Coverage of shared edited neurons by shared active routes.} To assess how closely the shared edited units in Section~\ref{subsec:directionality} correspond to shared active routes, we compute the coverage rate
\begin{equation}
\mathrm{Cov}(A,B) = \frac{ \sum_{\ell=1}^{L} \bigl| \mathcal{Q}_{A,B}^{(\ell)} \cap \mathcal{A}_{A}^{(\ell)} \cap \mathcal{A}_{B}^{(\ell)} \bigr| }{ \sum_{\ell=1}^{L} \bigl| \mathcal{Q}_{A,B}^{(\ell)} \bigr| },
\end{equation}
where $\mathcal{A}_{d}^{(\ell)}$ is the Top-10\% active-neuron set of Appendix~\ref{app:active_route_definition}. Averaged over all domain pairs and all layers, $\mathrm{Cov}=97.39\%$: almost every unit that two domains both edit heavily is also a unit that both domains activate heavily. This supports analyzing directional alignment on these units as a diagnostic of shared routes. The remaining $2.61\%$ lies outside the shared top-active intersection.

\section{Detailed Proof of Our Theory}
\label{app:det_proof}

\subsection{Structural Assumptions}
\label{app:theory_assumptions}

The analysis uses a fixed local objective $L_d$ for each domain. The diagonal operator $\mW$ defines the actual rollback, with $\vv=\mW\vdelta_B$ and $\vr=(\mI-\mW)\vdelta_B$. The orthogonal projector $\mP_S$ is used separately in the idealized refresh analysis. The following assumptions state the conditions of the bounds; we do not report empirical estimates of all their constants.

\paragraph{Assumption 1: local smoothness.}
Each $L_d$ is twice continuously differentiable on a neighborhood containing the relevant segments and iterates, with $\|\mH_d(\vtheta)\|_2\le\beta_d$. The path average
\begin{equation}
\bar\mH_A=2\int_0^1(1-s)\mH_A(\vtheta_A^*+s\vdelta_B)\,ds
\end{equation}
defines the exact curvature term in Propositions~1 and~2.

\paragraph{Assumption 2: approximate stationarity.}
At the reference point,
\begin{equation}
\|\vg_A(\vtheta_A^*)\|_2\le\varepsilon_A.
\end{equation}
This condition controls the first-order upper bound and the additive term in the refresh bound. The damage identities themselves retain the signed drift $a$ and do not require it to vanish.

\paragraph{Assumption 3: local refresh curvature, coupling, and radius.}
Let $\ve_t=\vtheta_t-\vtheta_A^*$ and $\tilde\mH_t=\int_0^1\mH_A(\vtheta_A^*+s\ve_t)\,ds$. Along the refresh trajectory, suppose $\|\ve_t\|_2\le R$ and
\begin{equation}
\mu_A\mI_S\preceq\mP_S\tilde\mH_t\mP_S\big|_{S_{A,B}}\preceq\bar\beta_A\mI_S,
\qquad
\|\mP_S\tilde\mH_t(\mI-\mP_S)\|_2\le\xi_A,
\end{equation}
where $0<\mu_A\le\bar\beta_A$. The step size satisfies $0<\alpha\le1/\bar\beta_A$. These are conditions on the relevant average curvature and cross-subspace coupling, not a claim of global convexity.

\subsection{Proof of Proposition 1}
\label{app:proof_second_order}

We provide the full derivation for Proposition~1 and make explicit its sensitivity-based interpretation. Recall that training first reaches a domain-$A$ checkpoint $\vtheta_A^*$ and later domain-$B$ training induces a local displacement $\vdelta_B$, so the degraded checkpoint is $\vtheta_A^*+\vdelta_B$. The interference from $B$ to $A$ is
\begin{equation}
\Delta_{A\leftarrow B} = L_A(\vtheta_A^*+\vdelta_B)-L_A(\vtheta_A^*).
\end{equation}
Let $\phi(s)=L_A(\vtheta_A^*+s\vdelta_B)$ for $s\in[0,1]$, which is twice continuously differentiable by Assumption~1 with $\phi'(s)=\vg_A(\vtheta_A^*+s\vdelta_B)^\top\vdelta_B$ and $\phi''(s)=\vdelta_B^\top\mH_A(\vtheta_A^*+s\vdelta_B)\vdelta_B$. Integrating by parts twice, or equivalently applying Taylor's theorem with integral remainder to $\phi$ at $s=0$,
\begin{equation}
\Delta_{A\leftarrow B} = \phi(1)-\phi(0) = \phi'(0) + \int_0^1 (1-s)\,\phi''(s)\,ds .
\end{equation}
Substituting $\phi'(0)=\vg_A(\vtheta_A^*)^\top\vdelta_B$ and $\phi''(s)=\vdelta_B^\top\mH_A(\vtheta_A^*+s\vdelta_B)\vdelta_B$ gives
\begin{equation}
\Delta_{A\leftarrow B} = \vg_A(\vtheta_A^*)^\top\vdelta_B + \int_0^1 (1-s)\,\vdelta_B^\top\mH_A(\vtheta_A^*+s\vdelta_B)\vdelta_B\,ds .
\end{equation}
By the definition of the path-averaged Hessian $\bar\mH_A=2\int_0^1(1-s)\mH_A(\vtheta_A^*+s\vdelta_B)\,ds$ and of the normalized path curvature $\bar\kappa_A(\vu)=\vu^\top\bar\mH_A\vu/\|\vdelta_B\|_2^2$, the second term equals $\tfrac12\vdelta_B^\top\bar\mH_A\vdelta_B=\tfrac12\|\vdelta_B\|_2^2\,\bar\kappa_A(\vdelta_B)$, so
\begin{equation}
\Delta_{A\leftarrow B} = \vg_A(\vtheta_A^*)^\top\vdelta_B + \tfrac12\|\vdelta_B\|_2^2\,\bar\kappa_A(\vdelta_B).
\label{eq:prop1_identity}
\end{equation}
This is the integral form of the loss change, with the first-order drift retained. Because the average Hessian depends on the displacement, the identity does not assert that loss scales quadratically over arbitrary displacement magnitudes.

Under Assumption~2 the checkpoint $\vtheta_A^*$ is approximately stationary for domain $A$, which bounds the first term by
\begin{equation}
|\vg_A(\vtheta_A^*)^\top \vdelta_B| \le \|\vg_A(\vtheta_A^*)\|_2\|\vdelta_B\|_2 \le \varepsilon_A\|\vdelta_B\|_2,
\end{equation}
and substituting into Equation~\ref{eq:prop1_identity} yields
\begin{equation}
\left| \Delta_{A\leftarrow B} - \tfrac{1}{2}\|\vdelta_B\|_2^2\,\bar\kappa_A(\vdelta_B) \right| \le \varepsilon_A\|\vdelta_B\|_2,
\label{eq:prop1_bound}
\end{equation}
which proves Proposition~1. The identity separates first-order drift from the curvature contribution without dropping a higher-order remainder, and it separates displacement magnitude $\|\vdelta_B\|_2^2$ from directional sensitivity $\bar\kappa_A(\vdelta_B)$. The norm-based bound on $a$ may be loose: neither a small directional term nor a small gradient cosine alone fixes the relative size of the curvature contribution, so curvature dominates only when its full contribution is large relative to $|a|$.

\subsection{Proof of Proposition 2}
\label{app:proof_conflict_subspace}

Since $\vdelta_B=\vv+\vr$ and $\bar\mH_A$ is symmetric,
\begin{equation}
\vdelta_B^\top\bar\mH_A\vdelta_B
=\vv^\top\bar\mH_A\vv+2\vv^\top\bar\mH_A\vr+\vr^\top\bar\mH_A\vr
=Q_v+2C+Q_r.
\label{eq:three_terms}
\end{equation}
Substitution into Proposition~1 gives Equation~\ref{eq:route_decomposition}. For $Q_v>0$, division by $Q_v/2$ and the triangle inequality yield
\begin{equation}
\left|\frac{\Delta_{A\leftarrow B}}{Q_v/2}-1\right|
=\left|\frac{2a+2C+Q_r}{Q_v}\right|
\le\frac{2|a|}{Q_v}+\frac{2|C|+|Q_r|}{Q_v},
\end{equation}
which proves Equation~\ref{eq:localization_certificate}. No orthogonality or idempotence of $\mW$ is required: all three quadratic forms use the same path-averaged Hessian. The decomposition retains the interaction between the reverted and retained components.

\subsection{Proof of Theorem 1}
\label{app:proof_refresh}

Let
\begin{equation}
\ve_t=\vtheta_t-\vtheta_A^*, \quad \ve_t^S=\mP_S\ve_t.
\end{equation}
The refresh iteration is
\begin{equation}
\vtheta_{t+1}=\vtheta_t-\alpha \vg_A(\vtheta_t),
\end{equation}
which implies
\begin{equation}
\ve_{t+1}=\ve_t-\alpha \vg_A(\vtheta_t).
\end{equation}
Projecting onto the shared conflict subspace gives
\begin{equation}
\ve_{t+1}^S = \mP_S\ve_t-\alpha \mP_S \vg_A(\vtheta_t) = \ve_t^S-\alpha \mP_S \vg_A(\vtheta_t).
\end{equation}

By the mean value theorem for vector-valued maps, which Assumption~1 licenses on the segment from $\vtheta_A^*$ to $\vtheta_t$, the gradient decomposes \textit{exactly} as
\begin{equation}
\vg_A(\vtheta_t) = \vg_A(\vtheta_A^*) + \tilde\mH_t\ve_t, \qquad \tilde\mH_t=\int_0^1\mH_A(\vtheta_A^*+s\ve_t)\,ds ,
\end{equation}
with no remainder, so no Hessian-Lipschitz constant and no quadratic error term enter the recursion. Substituting into the projected recursion gives
\begin{equation}
\ve_{t+1}^S ={} \ve_t^S - \alpha \mP_S \vg_A(\vtheta_A^*) - \alpha \mP_S \tilde\mH_t\ve_t.
\end{equation}
Decompose $\ve_t=\ve_t^S+\ve_t^{\perp}$ with $\ve_t^{\perp}=(\mI-\mP_S)\ve_t$. Then
\begin{equation}
\mP_S \tilde\mH_t\ve_t ={} \mP_S \tilde\mH_t\mP_S \ve_t + \mP_S \tilde\mH_t(\mI-\mP_S)\ve_t ,
\end{equation}
and the projected recursion becomes
\begin{equation}
\ve_{t+1}^S ={} \left(\mI- \alpha \mP_S \tilde\mH_t\mP_S\right)\ve_t^S - \alpha \mP_S \vg_A(\vtheta_A^*) - \alpha\mP_S \tilde\mH_t(\mI-\mP_S)\ve_t .
\label{eq:projected_recursion}
\end{equation}
Assumptions~2 and~3 bound the three terms on the right in turn. First, since $\mu_A \mI_S\preceq \mP_S\tilde\mH_t\mP_S\vert_{S_{A,B}}\preceq \bar\beta_A \mI_S$ and $0<\alpha\le 1/\bar\beta_A$, every eigenvalue of $\mI-\alpha \mP_S \tilde\mH_t\mP_S$ restricted to $S_{A,B}$ lies in $[0,1-\alpha\mu_A]$, so as an operator on $S_{A,B}$
\begin{equation}
\bigl\|\mI-\alpha \mP_S \tilde\mH_t\mP_S\bigr\|_{2,S} \le 1-\alpha\mu_A .
\end{equation}
Second, by Assumption~2, $\|\mP_S \vg_A(\vtheta_A^*)\|_2\le \|\vg_A(\vtheta_A^*)\|_2\le \varepsilon_A$. Third, by the coupling bound of Assumption~3 together with the trust region $\|\ve_t\|_2\le R$, which controls $\|\ve_t^\perp\|_2\le R$,
\begin{equation}
\bigl\|\mP_S \tilde\mH_t(\mI-\mP_S)\ve_t\bigr\|_2 \le \xi_A\|\ve_t^{\perp}\|_2 \le \xi_A R .
\end{equation}
Taking norms in Equation~\ref{eq:projected_recursion} therefore yields a recursion in which the perturbation is a constant rather than a function of the iterate being bounded:
\begin{equation}
\|\ve_{t+1}^S\|_2 \le (1-\alpha\mu_A)\|\ve_t^S\|_2 + \alpha\bigl(\varepsilon_A+\xi_A R\bigr).
\end{equation}
Unrolling and using $\ve_0=\vdelta_B$, so that $\ve_0^S=\mP_S\vdelta_B$, together with $\sum_{k=0}^{t-1}(1-\alpha\mu_A)^{k}\le 1/(\alpha\mu_A)$, gives
\begin{equation}
\|\mP_S(\vtheta_t-\vtheta_A^*)\|_2 \le (1-\alpha\mu_A)^t\|\mP_S\vdelta_B\|_2 + \frac{\varepsilon_A+\xi_A R}{\mu_A},
\end{equation}
which proves the theorem.

The additive term $(\varepsilon_A+\xi_AR)/\mu_A$ is an allowance in the upper bound, not a residual the iterate must incur: actual projected errors can be smaller, including convergence to zero when the allowance is positive.

\begin{table*}[t]
\centering
\scriptsize
\caption{Full benchmark-level results; Step is the selected checkpoint step and AVG averages the four evaluation domains (Math, Code, QA, CW).}
\label{tab:full_task_results}
\resizebox{\textwidth}{!}{
\begin{tabular}{lccccccccccccc}
\toprule
\multirow{2}{*}{Model} & \multirow{2}{*}{Step}
& \multicolumn{6}{c}{Math}
& \multicolumn{1}{c}{Code}
& \multicolumn{3}{c}{QA}
& \multicolumn{1}{c}{CW}
& \multirow{2}{*}{AVG} \\
\cmidrule(lr){3-8}\cmidrule(lr){9-9}\cmidrule(lr){10-12}\cmidrule(lr){13-13}
&
& AIME24 & AIME25 & AIME26 & OlympiadBench & HMMT & Math
& LiveCodeBench-v6
& SuperGPQA-test & MMLU-Pro & QA
& WritingBench & \\
\midrule
Base & --
& 48.75 & 42.50 & 40.63 & 61.68 & 22.40 & 43.19
& 29.57
& 47.32 & 73.96 & 60.64
& 82.44 & 53.96 \\

$\mathrm{Math}_s$ & 525
& 71.77 & 67.71 & 69.79 & 77.95 & 46.98 & 66.84
& 34.65
& 47.84 & 73.68 & 60.76
& 81.38 & 60.91 \\

$\mathrm{Code}_s$ & 600
& 61.46 & 61.46 & 62.71 & 71.38 & 41.15 & 59.63
& 52.67
& 47.84 & 73.93 & 60.89
& 82.40 & 63.90 \\

$\mathrm{QA}_s$ & 705
& 58.65 & 55.83 & 56.77 & 70.19 & 35.10 & 55.31
& 32.07
& 51.85 & 74.77 & 63.31
& 81.76 & 58.11 \\

$\mathrm{CW}_s$ & 120
& 44.27 & 38.85 & 35.00 & 59.74 & 21.04 & 39.78
& 28.15
& 47.16 & 73.84 & 60.50
& 86.24 & 53.67 \\

$\mathrm{Code}_o$ & 600
& 61.46 & 61.46 & 62.71 & 71.38 & 41.15 & 59.63
& 52.67
& 47.84 & 73.93 & 60.89
& 82.40 & 63.90 \\

$\mathrm{Math}_o$ & 180
& 73.33 & 66.56 & 70.83 & 77.57 & 44.17 & 66.49
& 50.69
& 47.53 & 73.51 & 60.52
& 81.44 & 64.79 \\

$\mathrm{QA}_o$ & 300
& 64.38 & 59.17 & 61.98 & 72.73 & 41.25 & 59.90
& 50.99
& 49.98 & 74.69 & 62.34
& 81.79 & 63.76 \\

$\mathrm{CW}_o$ & 135
& 62.29 & 56.04 & 58.65 & 70.89 & 40.42 & 57.66
& 50.47
& 49.91 & 74.77 & 62.34
& 86.52 & 64.25 \\

CGPO & 630
& 65.10 & 58.13 & 66.88 & 75.35 & 44.17 & 61.93
& 50.05
& 50.90 & 74.05 & 62.48
& 86.73 & 65.30 \\

JT & 1035
& 69.58 & 65.63 & 65.52 & 77.12 & 46.15 & 64.80
& 48.61
& 50.04 & 74.18 & 62.11
& 86.97 & 65.62 \\

Replay-all & 150
& 68.85 & 62.19 & 66.88 & 74.99 & 44.79 & 63.54
& 51.09
& 50.43 & 74.69 & 62.56
& 86.08 & 65.82 \\

Re-Math & 135
& 74.27 & 64.48 & 68.65 & 77.06 & 45.73 & 66.04
& 51.05
& 50.28 & 74.69 & 62.49
& 85.96 & 66.39 \\

QA $\rightarrow$ Math & 675
& 71.15 & 66.77 & 68.54 & 77.04 & 43.13 & 65.33
& 32.53
& 51.55 & 74.44 & 63.00
& 82.01 & 60.72 \\

Math $\rightarrow$ CW & 120
& 72.71 & 67.71 & 70.42 & 77.75 & 47.19 & 67.16
& 32.82
& 48.31 & 74.04 & 61.18
& 85.77 & 61.73 \\
\bottomrule
\end{tabular}
}
\end{table*}

\subsection{Proof of Proposition 3}
\label{app:proof_other_domains}

Let $\vr_T=\vtheta_T-\vtheta_0$. Smoothness and $\|\vg_C(\vtheta_0)\|_2\le\varepsilon_C$ give
\begin{equation}
L_C(\vtheta_T)-L_C(\vtheta_0)
\le\vg_C(\vtheta_0)^\top\vr_T+\tfrac12\beta_C\|\vr_T\|_2^2
\le\varepsilon_C\|\vr_T\|_2+\tfrac12\beta_C\|\vr_T\|_2^2.
\end{equation}
For the gradient updates considered in Theorem~1,
\begin{equation}
\|\vr_T\|_2=\left\|-\alpha\sum_{t<T}\vg_A(\vtheta_t)\right\|_2
\le\alpha\sum_{t<T}\|\vg_A(\vtheta_t)\|_2\le\alpha TG_A,
\label{eq:displacement_in_T}
\end{equation}
where $G_A=\sup_{t<T}\|\vg_A(\vtheta_t)\|_2$. Substitution proves Equation~\ref{eq:collateral_bound}. The bound depends on displacement and local sensitivity; its size is not determined by the refresh length alone.

\begin{figure*}[t]
  \centering
  \includegraphics[width=0.95\textwidth]{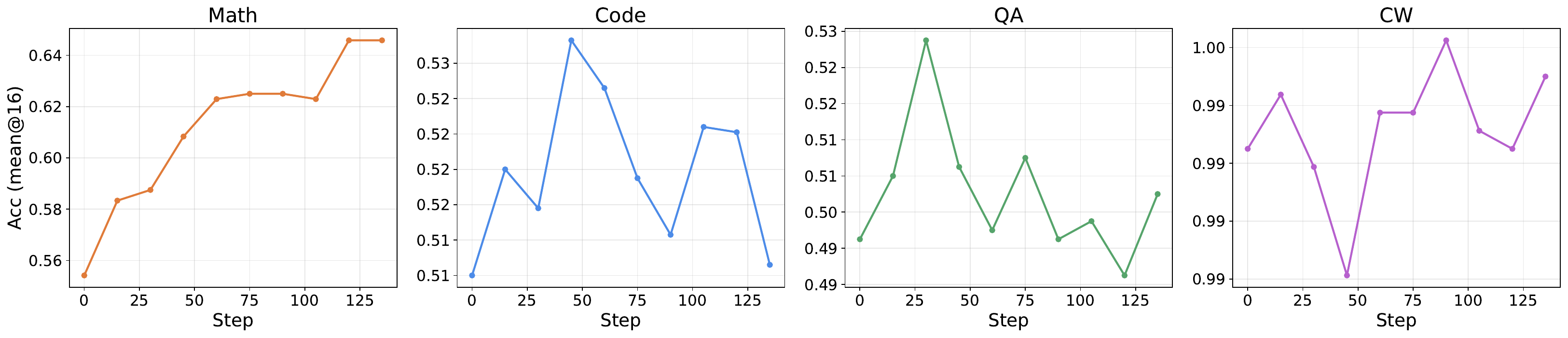}
  \caption{Validation dynamics during Re-Math refresh from $\mathrm{CW}_o$ across the four domains.}
  \label{fig:refresh_dynamics}
\end{figure*}

\subsection{Extension: Alternating Refresh as Local Multi-Objective Optimization}
\label{app:proof_alt_refresh}

A short alternating-refresh cycle has a useful first-order interpretation as a weighted local update. Consider two domains $A$ and $B$. One cycle first updates on $A$:
\begin{equation}
\vtheta' = \vtheta-\alpha_A \vg_A(\vtheta),
\end{equation}
and then updates on $B$:
\begin{equation}
\vtheta'' = \vtheta'-\alpha_B \vg_B(\vtheta').
\end{equation}
For small step sizes, local smoothness gives
\begin{equation}
\vg_B(\vtheta') = \vg_B(\vtheta)+O(\alpha_A\|\vg_A(\vtheta)\|_2).
\end{equation}
Thus, one refresh cycle satisfies
\begin{equation}
\vtheta'' = \vtheta - \alpha_A \vg_A(\vtheta) - \alpha_B \vg_B(\vtheta) + O(\alpha_A\alpha_B).
\end{equation}
To first order, alternating refresh therefore follows gradient descent on the weighted local objective
\begin{equation}
\Phi(\vtheta) = \alpha_A L_A(\vtheta) + \alpha_B L_B(\vtheta).
\end{equation}
More generally, alternating refresh over multiple domains approximates gradient descent on
\begin{equation}
\Phi(\vtheta) = \sum_{d\in\mathcal{D}} \alpha_d L_d(\vtheta).
\end{equation}

A stationary point of this local objective satisfies
\begin{equation}
\sum_{d\in\mathcal{D}} \alpha_d \vg_d(\vtheta^\dagger) \approx 0,
\end{equation}
which for positive weights $\{\alpha_d\}$ is the stationarity condition of the weighted objective. The cycle approximation motivates viewing alternating refresh as balancing local domain objectives; it does not establish convergence to that point or a Pareto improvement.

\section{Additional Validation Results}
\subsection{Full Task-Level Results}
\label{app:full_results}

Table~\ref{tab:full_task_results} reports the full benchmark-level results for all model checkpoints used in the main task-level validation. For sequential and refresh runs, the step number in parentheses denotes the selected checkpoint step within the last training stage.

Because AIME25 doubles as the Math validation set (Section~\ref{sec:setup}), we also recompute the Math average over the four held-out Math benchmarks. The pattern is unchanged: Math is $66.48$ at $\mathrm{Math}_o$, falls to $60.09$ at $\mathrm{QA}_o$ and $58.06$ at $\mathrm{CW}_o$, and returns to $66.43$ after Re-Math, versus $63.88$ for Replay-all and $64.59$ for $\mathrm{JT}$. The QA-induced drop measured this way is $6.39$ points instead of $6.59$, and the ordering of all settings is preserved.

\subsection{Refresh Dynamics}
\label{app:refresh_dynamics}

\begin{wrapfigure}{r}{0.43\textwidth}
  \vspace{-3.0em}
  \centering
  \includegraphics[width=0.42\textwidth]{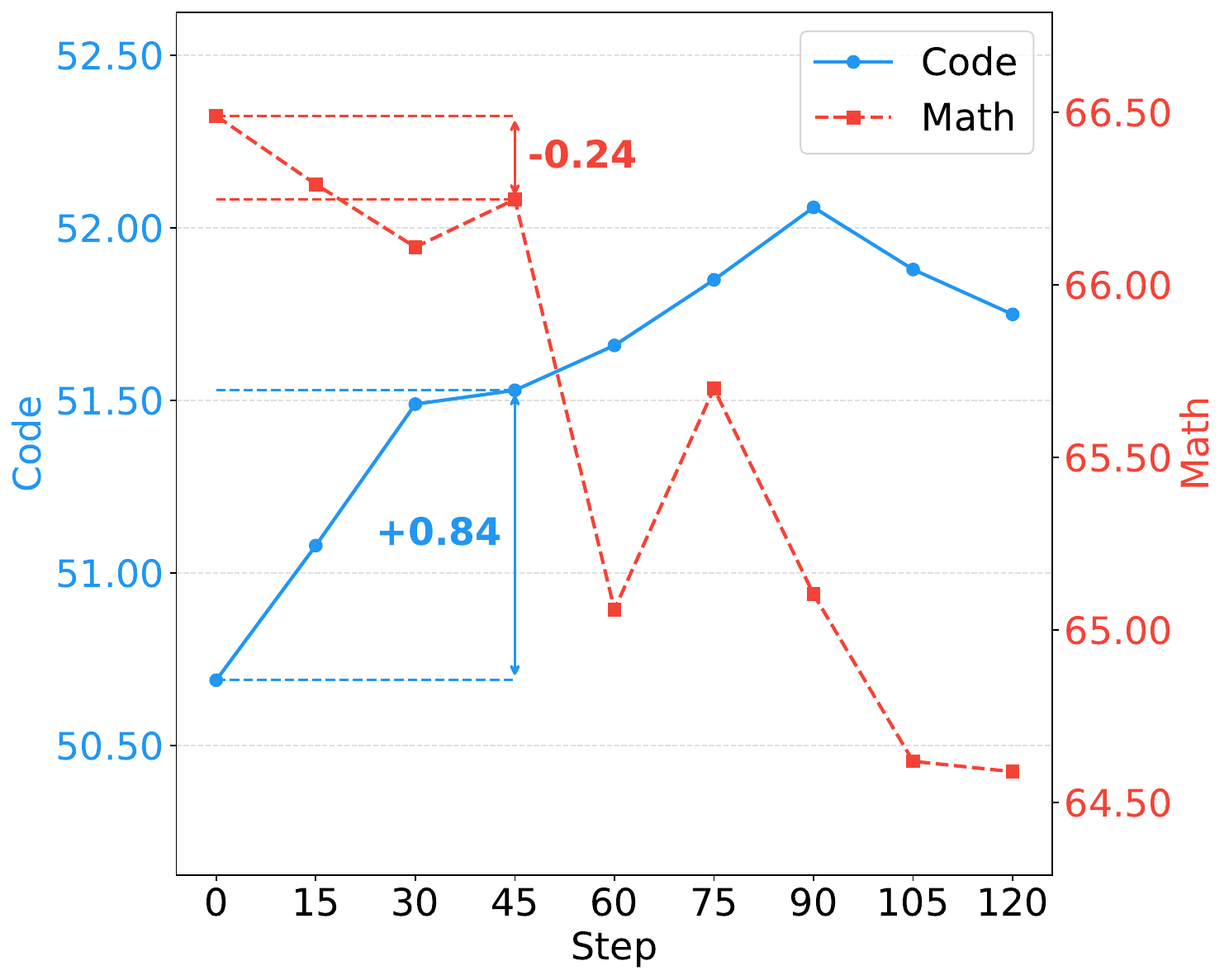}
  \caption{Math and Code performance changes during Re-Code on $\mathrm{Math}_o$.}
  \label{fig:recode_dual_axis}
  \vspace{-2.0em}
\end{wrapfigure}

Figure~\ref{fig:refresh_dynamics} shows rapid Math improvement early in Re-Math, followed by saturation, while the other domains change comparatively little. This behavior is consistent with local repair and the conditional interpretation of Theorem~1; the accuracy curves are not measurements of its parameter-space contraction bound.

Figure~\ref{fig:recode_dual_axis} provides a complementary Code $\rightarrow$ Math $\rightarrow$ Re-Code example. Early refresh improves Code with modest Math changes, while longer specialization produces a more visible Math decline. This supports using a limited refresh budget and is consistent with displacement-dependent sensitivity in Proposition~1.

\subsection{Directional Asymmetry of Interference}
\label{app:asym_directionality}

\begin{wrapfigure}{r}{0.50\textwidth}
  \vspace{-1.5em}
  \centering
  \includegraphics[width=0.48\textwidth]{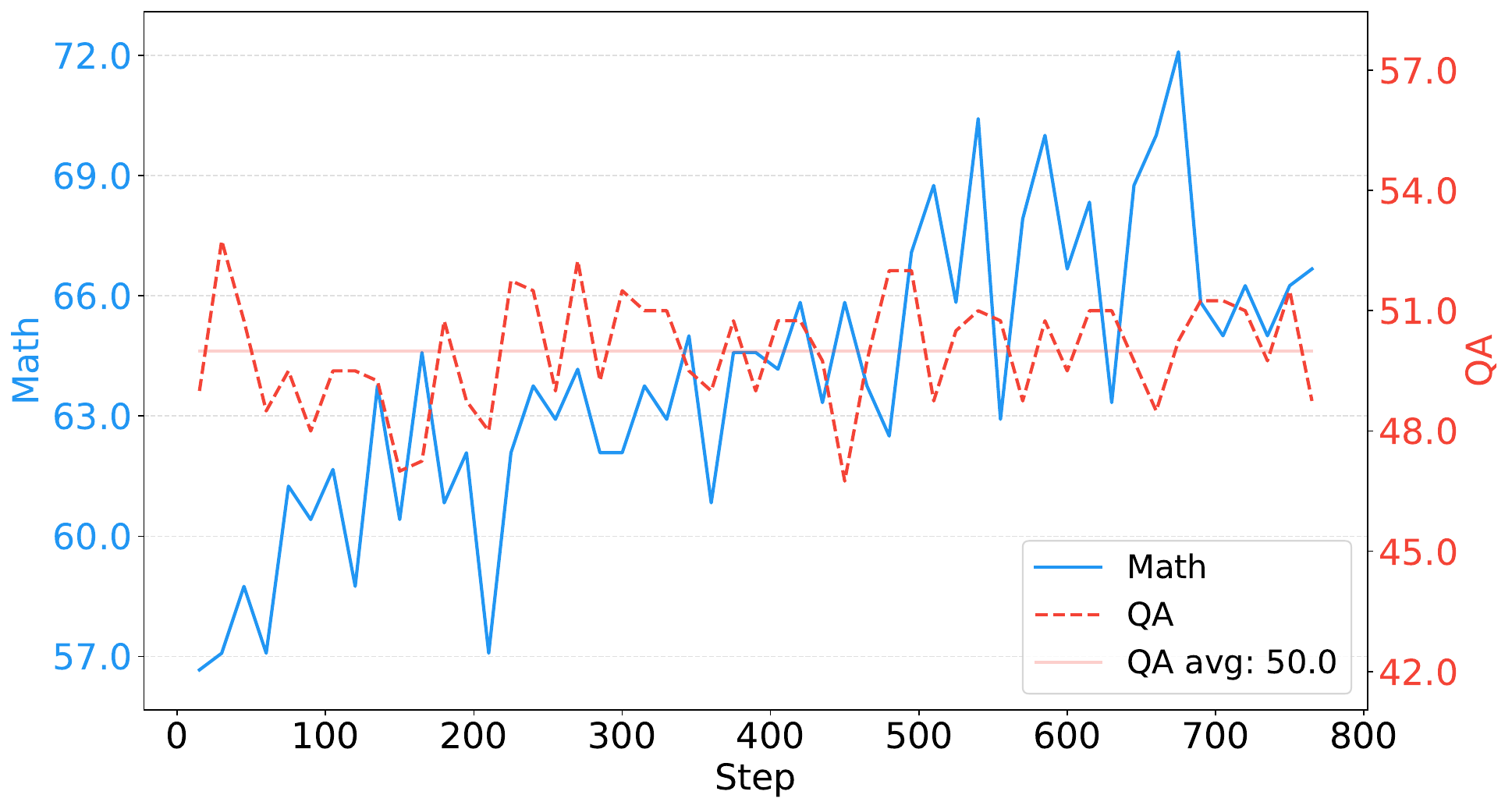}
  \caption{Validation dynamics for the reverse QA $\rightarrow$ Math ordering.}
  \label{fig:qa_math_dynamics}
  \vspace{-3.0em}
\end{wrapfigure}

We additionally compare QA $\rightarrow$ Math with the reverse Math $\rightarrow$ QA ordering. As shown in Figure~\ref{fig:qa_math_dynamics}, in QA $\rightarrow$ Math, Math improves steadily during the Math stage while QA remains largely stable, unlike the stronger Math degradation observed when QA is trained after Math.

This provides further evidence that cross-domain interference is directional rather than symmetric. This asymmetry is also consistent with our local perturbation analysis and Proposition~1: the damage to an earlier domain depends on whether the later update enters its curvature-sensitive shared directions. Here, the Math update appears to perturb QA-sensitive directions only weakly, whereas the QA update more strongly perturbs Math-sensitive directions in the reverse ordering.

\subsection{Generalization to Another Model Family}
\label{app:cross_family}

To test that our findings are not specific to the Qwen3-4B-Thinking model used in the main text, we replicate the core sequential-interference-and-refresh phenomenon on a Phi-4-mini-reasoning model under a Math~$\rightarrow$~QA~$\rightarrow$~Re-Math schedule (Table~\ref{tab:cross_family}). RL on Math first lifts both Math ($40.15\rightarrow48.14$) and QA ($38.94\rightarrow47.27$). Subsequent QA training then improves QA to $53.40$ while degrading Math to $47.56$---a smaller but directionally consistent drop---reproducing the asymmetric cross-domain interference of Section~\ref{sec:ana}. A short $45$-step Re-Math refresh recovers Math to its best value ($48.76$, above the earlier Math peak) while leaving QA essentially unchanged ($53.40\rightarrow53.29$). The recovery with little change in QA extends the task-level pattern across model families.

\begin{table}[t]
\centering
\scriptsize
\setlength{\tabcolsep}{3.2pt}
\caption{Cross-family results on a Phi-4-mini-reasoning model; parentheses indicate selected checkpoint steps.}
\label{tab:cross_family}
\resizebox{\textwidth}{!}{%
\begin{tabular}{lccccccccc}
\toprule
& \multicolumn{6}{c}{Math} & \multicolumn{3}{c}{QA} \\
\cmidrule(lr){2-7} \cmidrule(lr){8-10}
Checkpoint & AIME24 & AIME25 & AIME26 & OlyBench & HMMT & Avg. & SuperGPQA & MMLU-Pro & Avg. \\
\midrule
Base & 46.46 & 31.88 & 37.40 & 63.32 & 21.67 & 40.15 & 26.12 & 51.76 & 38.94 \\
$+$ Math (1065) & 55.94 & 44.58 & 45.73 & 69.43 & 25.00 & 48.14 & 32.74 & 61.79 & 47.27 \\
$+$ QA (390) & 55.42 & 43.02 & 45.00 & 69.69 & 24.69 & 47.56 & 38.25 & 68.54 & \textbf{53.40} \\
$+$ Re-Math (45) & \textbf{57.60} & \textbf{44.79} & \textbf{46.67} & \textbf{69.72} & 25.00 & \textbf{48.76} & 38.09 & 68.48 & 53.29 \\
\bottomrule
\end{tabular}
}
\end{table}

\subsection{Direct Rollback on a Coordinate Proxy for the Conflict Subspace}
\label{app:proxy_conflict_appendix}

To more directly probe the localization claim behind Proposition~2, we intervene on the checkpoint pair $\mathrm{Code}_o \rightarrow \mathrm{Math}_o \rightarrow \mathrm{QA}_o$.
Treating checkpoints as their parameter vectors, define the QA-induced displacement as
\begin{equation}
\vdelta_{Q\mid M}=\mathrm{QA}_o-\mathrm{Math}_o.
\end{equation}
We restrict the intervention to MLP neurons, using the same neuron definition as in Appendix~\ref{app:neuron_definition}. For neuron $i$ in layer $\ell$, let $\Delta \vv_{M,i}^{(\ell)}$ and $\Delta \vv_{Q,i}^{(\ell)}$ denote the Math and QA task vectors in the concatenated gate/up/down parameterization from Appendix~\ref{app:directional_alignment_metric}.

We build a neuron-level proxy score
\begin{equation}
S_i^{(\ell)} = A_i^{(\ell)} M_i^{(\ell)} C_i^{(\ell)},
\end{equation}
where
\begin{equation}
A_i^{(\ell)}= \min\!\Big( \operatorname{Pct}(a_{M,i}^{(\ell)}), \operatorname{Pct}(a_{Q,i}^{(\ell)}) \Big),
\end{equation}
\begin{equation}
M_i^{(\ell)}= \operatorname{Pct}\!\left(\|\Delta \vv_{Q,i}^{(\ell)}\|_2\right),
\end{equation}
and
\begin{equation}
C_i^{(\ell)}= \max\!\left( 0,\, -\cos\!\left(\Delta \vv_{M,i}^{(\ell)},\Delta \vv_{Q,i}^{(\ell)}\right) \right).
\end{equation}
Here $a_{d,i}^{(\ell)}$ is the dataset-level activation score defined in Appendix~\ref{app:active_route_definition}, and $\operatorname{Pct}(\cdot)$ denotes percentile rank within a layer.

For activation collection, we use 512 samples and first generate up to 4096 tokens, followed by masked teacher-forcing forward passes to record MLP intermediate activations. Unless otherwise stated, the MLP-only selector uses a budget of $1.3\%$, and we write $B$ for the corresponding number of selected units.

To distribute the global budget across layers, we first compute a layer score by averaging the top-$\rho$ neuron scores within each layer, with $\rho=10\%$. The final layer budget is then allocated proportionally to
\begin{equation}
\exp\!\left((\bar S^{(\ell)})^{\gamma}\right),
\end{equation}
with a sharpening exponent $\gamma=2.0$, followed by rounding so that $\sum_{\ell} B_\ell=B$.

Given the selected neuron set $\hat S$, we revert only the QA increment on its gate/up/down parameters:
\begin{equation}
\vtheta_{\mathrm{rev}} = \mathrm{QA}_o-\mP_{\hat S}\vdelta_{Q\mid M},
\end{equation}
where $\mP_{\hat S}$ is the coordinate projection induced by $\hat S$. This intervention requires no additional optimization and provides a direct proxy-level test of whether a sparse subset of the QA displacement carries measurable Math damage.

For single-factor selectors (A, M, or C), we report two variants. In the fixed-budget variant, we keep the layer allocation from the full $A\times M\times C$ selector and only change the within-layer ranking. In the recomputed-budget variant, we recompute the layer allocation from the corresponding selector itself.

\begin{table*}[t]
\centering
\scriptsize
\setlength{\tabcolsep}{3.2pt}
\caption{Additional ablations for the proxy conflict-subspace intervention at a $1.3\%$ budget: fixed-budget variants keep the full selector's layer allocation and change only within-layer ranking, while recomputed-budget variants recompute both ranking and layer allocation.}
\label{tab:proxy_conflict_ablation_full}
\resizebox{\textwidth}{!}{%
\begin{tabular}{lccccc|ccc|cccc}
\toprule
Selector & AIME24 & AIME25 & AIME26 & OlyBench & HMMT & Math Avg & $\Delta$ vs $\mathrm{QA}_o$ & Recovery & SuperGPQA & MMLU-Pro & QA Avg & $\Delta$ QA Avg \\
\midrule
$\mathrm{Math}_o$ & 73.33 & 66.56 & 70.83 & 77.57 & 44.17 & 66.49 & +6.59 & 100.0\% & 47.53 & 73.51 & 60.52 & $-1.81$ \\
$\mathrm{QA}_o$ & 64.38 & 59.17 & 61.98 & 72.73 & 41.25 & 59.90 & 0.00 & 0.0\% & 49.98 & 74.69 & 62.34 & 0.00 \\
Random & 63.54 & 59.90 & 60.00 & 73.05 & 40.94 & 59.49 & $-0.41$ & $-6.3\%$ & 49.97 & 74.71 & 62.34 & +0.01 \\
\midrule
A$\times$M$\times$C (main) & 66.25 & 60.21 & 63.65 & 73.42 & 42.71 & 61.25 & +1.35 & \textbf{20.4\%} & 50.11 & 74.43 & 62.27 & $-0.06$ \\
M$\times$C & 66.67 & 60.00 & 61.77 & 73.36 & 43.75 & 61.11 & +1.21 & 18.3\% & 50.16 & 74.58 & 62.37 & +0.04 \\
A$\times$C & 65.00 & 59.48 & 62.81 & 73.50 & 41.98 & 60.55 & +0.65 & 9.9\% & 49.92 & 74.65 & 62.29 & $-0.05$ \\
A$\times$M & 65.73 & 61.46 & 61.77 & 72.97 & 40.73 & 60.53 & +0.63 & 9.6\% & 50.25 & 74.49 & 62.37 & +0.04 \\
M (fixed budget) & 64.69 & 60.10 & 63.12 & 73.50 & 42.40 & 60.76 & +0.86 & 13.1\% & 49.93 & 74.56 & 62.25 & $-0.09$ \\
A (fixed budget) & 65.52 & 59.90 & 63.75 & 73.23 & 43.85 & 61.25 & +1.35 & \textbf{20.4\%} & 49.93 & 74.55 & 62.24 & $-0.09$ \\
C (fixed budget) & 65.83 & 60.21 & 61.77 & 73.30 & 42.60 & 60.74 & +0.84 & 12.7\% & 49.94 & 74.59 & 62.27 & $-0.07$ \\
M (recomputed) & 65.21 & 59.38 & 63.33 & 73.47 & 42.71 & 60.82 & +0.92 & 13.9\% & 50.14 & 74.69 & 62.41 & +0.08 \\
A (recomputed) & 66.15 & 60.42 & 61.88 & 73.41 & 43.12 & 61.00 & +1.09 & 16.6\% & 50.18 & 74.56 & 62.37 & +0.04 \\
C (recomputed) & 65.52 & 61.35 & 62.71 & 73.23 & 42.08 & 60.98 & +1.08 & 16.3\% & 50.15 & 74.57 & 62.36 & +0.03 \\
\bottomrule
\end{tabular}
}
\end{table*}

Table~\ref{tab:proxy_conflict_ablation_full} shows that the full selector recovers $20.4\%$ of the QA-induced Math loss, while the two-factor selector $M\times C$ remains close at $18.3\%$. By contrast, removing either update magnitude or directional conflict reduces recovery to about half of the full result. Interestingly, A-only under the full selector's layer budget matches the full selector, whereas A-only with its own recomputed budget drops to $16.6\%$. This indicates that shared activation alone is not sufficient to explain the best-performing intervention; part of its apparent strength comes from inheriting the conflict-aware layer allocation.

\begin{figure}[t]
  \centering
  \includegraphics[width=0.7\textwidth]{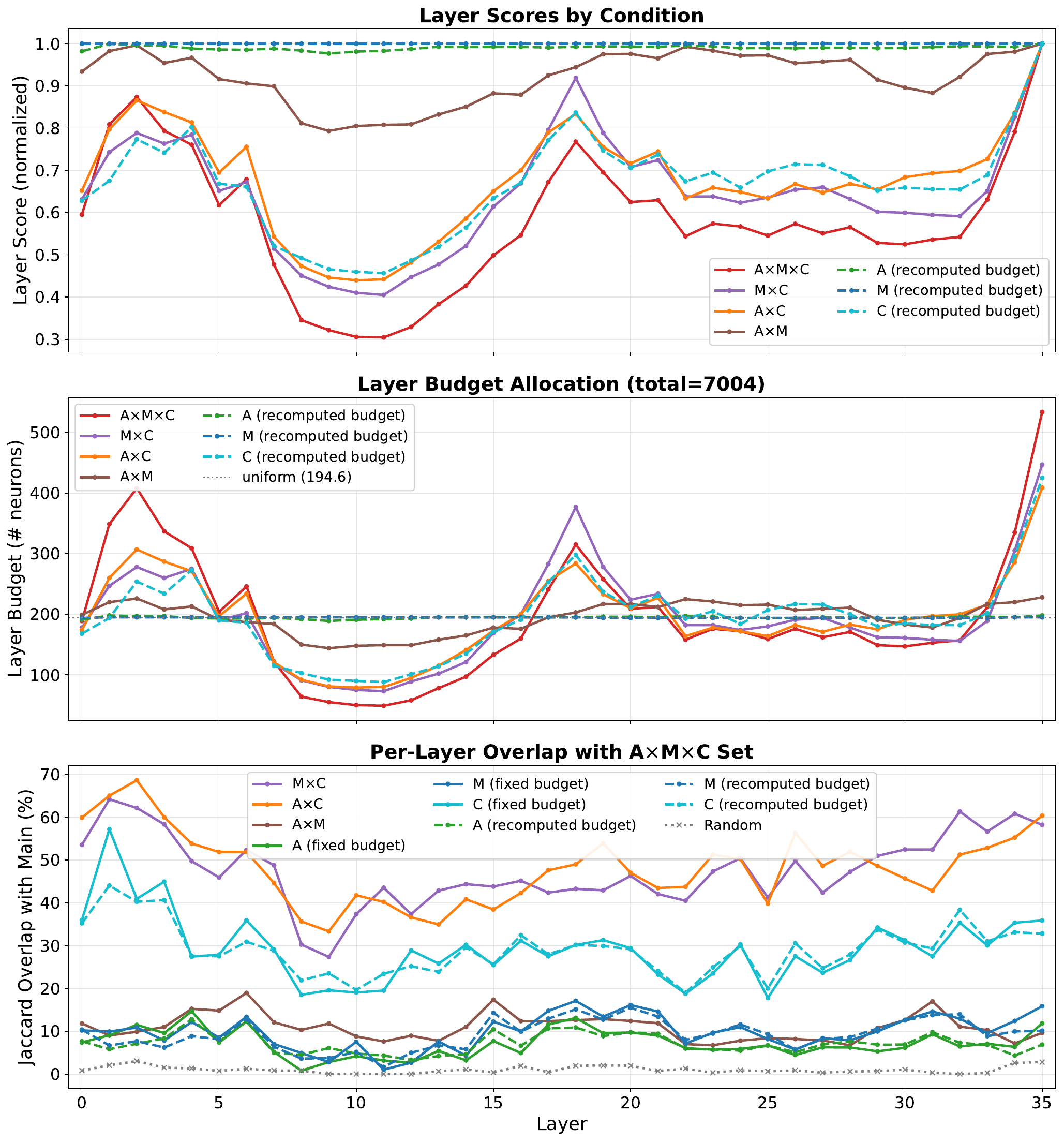}
  \caption{Layer-wise neuron selection across 36 layers---normalized layer scores (top), budget allocation (middle), and Jaccard overlap with the $A \times M \times C$ set (bottom)---where the conflict factor $C$ dominates the non-uniform budget and two-factor combinations containing $C$ recover the highest overlap.}
  \label{fig:proxy_conflict_budgets}
\end{figure}

Figure~\ref{fig:proxy_conflict_budgets} shows that the depth profile of the intervention is not uniform. Once the conflict term is included, the layer budget becomes strongly U-shaped, concentrating more neurons in shallow and late layers; without the conflict term, the budget becomes much flatter. The selector $M\times C$ is the closest two-factor approximation to the full selector, whereas A-only can exhibit much smaller support overlap despite competitive recovery in the fixed-budget setting. This suggests that the proxy support is partly redundant: different sparse coordinate subsets can play similar functional roles as long as they target the harmful QA displacement.

The MLP-only results above show that a sparse rollback on proxy conflict coordinates can already recover a substantial part of the QA-induced Math loss, but the recovery saturates early and remains incomplete. To test whether the recoverable harmful displacement also extends beyond MLP coordinates, we extend the intervention to attention layers.

For layer $\ell$, query head $h$, and row index $r$, we define a fine-grained attention unit
\begin{equation}
\resizebox{0.93\linewidth}{!}{$
\mathrm{AttnUnit}(\ell,h,r)=\{ \Delta W_Q^{(\ell)}[h,r,:],\; \Delta W_O^{(\ell)}[:,h,r], \frac{1}{\sqrt{g}}\Delta W_K^{(\ell)}[\kappa(h),r,:], \frac{1}{\sqrt{g}}\Delta W_V^{(\ell)}[\kappa(h),r,:] \}.
$}
\end{equation}
where $\kappa(h)=\lfloor h/g \rfloor$ maps a query head to its shared KV head under grouped-query attention, and $g=4$ is the group size. Since each layer has $32$ query heads and head dimension $128$, this gives $32\times 128=4096$ attention units per layer.

Unlike the MLP case, collecting per-unit activation statistics for attention would require decomposing attention outputs at the head-row level. We therefore use a conservative two-factor attention score
\begin{equation}
S_{\mathrm{attn}}^{(\ell,h,r)}= M_{\mathrm{attn}}^{(\ell,h,r)} \,C_{\mathrm{attn}}^{(\ell,h,r)},
\end{equation}
where $M_{\mathrm{attn}}$ measures the QA update magnitude of the unit and $C_{\mathrm{attn}}$ measures directional conflict between the Math and QA task vectors. MLP neurons keep the original three-factor score
\begin{equation}
S_{\mathrm{mlp}}^{(\ell,i)}= A_i^{(\ell)}M_i^{(\ell)}C_i^{(\ell)}.
\end{equation}

For the joint MLP+Attn selector we keep the same budget convention, letting $B$ denote the total number of selected units at a given budget. We first compute a joint layer score by combining the top-$\rho$ scores from MLP and attention units:
\begin{equation}
s_\ell= \frac{ \bar S_{\mathrm{mlp}}^{\rho}(\ell)\, d_{\mathrm{int}} + \bar S_{\mathrm{attn}}^{\rho}(\ell)\, n_{\mathrm{attn}} }{ d_{\mathrm{int}}+n_{\mathrm{attn}} },
\end{equation}
with $\rho=0.1$, $d_{\mathrm{int}}=9728$, and $n_{\mathrm{attn}}=4096$. The layer budget is then allocated by
\begin{equation}
b_\ell = B\cdot \frac{(s_\ell)^{\gamma}}{\sum_{\ell'} (s_{\ell'})^{\gamma}}, \quad \gamma=2.0.
\end{equation}
Within each layer, we merge all MLP-neuron and attention-unit scores into a single ranked list and select the top-$b_\ell$ units regardless of type.

For selected MLP neurons, the revert operation is identical to the MLP-only setting. For a selected attention unit $\mathrm{AttnUnit}(\ell,h,r)$, we fully revert the corresponding $W_Q$ and $W_O$ slices. For shared KV weights, if $k$ out of the $g$ query heads attached to the same KV head select row $r$, we revert the corresponding $W_K$ and $W_V$ rows by fraction $k/g$.

At a $9.8\%$ budget the joint selector recovers $57.4\%$ of the QA-induced Math accuracy drop (Table~\ref{tab:proxy_conflict_revert}), above the MLP-only result at a comparable budget. This supports including attention coordinates as additional repair targets. Recovery remains incomplete, and the accuracy measurements do not identify the location or curvature contribution of the remaining damage. Improving the selection of repair coordinates is a direction for future work.

\subsection{Curvature Measurement Details}
\label{app:curvature_details}

This section describes the directional curvature measurements in Section~\ref{sec:curvature_validation}.

\paragraph{Probe directions.} The proxy uses the $9.8\%$ joint MLP+Attn selection and its actual revert operator $\mW$. The direction $\vdelta_S=\mW\vdelta_{Q\mid M}$, with $\vdelta_{Q\mid M}=\mathrm{QA}_o-\mathrm{Math}_o$, is held fixed across checkpoints. Random controls use the same revert scheme: $R_{\mathrm{l}}$ matches per-layer, per-module unit counts, and $R_{\mathrm{g}}$ matches total MLP and attention counts. The squared direction norms are $0.0577$ for $\hat S$, $0.0482$ for $R_{\mathrm{l}}$, and $0.0495$ for $R_{\mathrm{g}}$. The reported curvature divides by each direction's own squared norm.

\paragraph{Finite-difference estimator.} The checkpoint-level probes of Section~\ref{sec:curvature_validation} compare directions at four points of the QA trajectory, so we need an estimator that is cheap enough to repeat across checkpoints and directions. We estimate the normalized curvature $\kappa(\hat S;\vtheta)$ by a centered finite difference of Math GRPO gradients:
\begin{equation}
\kappa(\hat S;\vtheta)
\approx
\frac{\vdelta_S^\top\!\left[\vg_{\mathrm{Math}}(\vtheta+h\vdelta_S)-\vg_{\mathrm{Math}}(\vtheta-h\vdelta_S)\right]}{2h\,\|\vdelta_S\|_2^2},
\end{equation}
where $h=\eta/\|\vdelta_S\|_2$ normalizes the perturbation to a fixed step size $\eta$ in parameter space. Gradients are computed in fp32 on a fixed batch of Math GRPO data with identical rollouts and rewards across the paired forward-backward passes, so that the difference isolates the curvature term. Unless otherwise stated, we use $\eta=10^{-3}$.

\paragraph{Checkpoints.} We probe four saved checkpoints: $\mathrm{Math}_o$, QA step 150, $\mathrm{QA}_o$ at step 300, and QA step 450. At each checkpoint, we use the same fixed rollback direction and matched controls to measure local directional sensitivity along the QA training trajectory.

\begin{table}[t]
\small
\centering
\caption{Sensitivity of the curvature estimate at $\mathrm{Math}_o$ to the finite-difference step size $\eta$.}
\label{tab:curvature_fd_sensitivity}
\begin{tabular}{lccc}
\toprule
$\eta$ & $\kappa(\hat S)$ at $\mathrm{Math}_o$ & $\nicefrac{\hat S}{R_{\mathrm{l}}}$ & $\nicefrac{\hat S}{R_{\mathrm{g}}}$ \\
\midrule
$3\times 10^{-4}$ & $1.58\times 10^{-3}$ & 3.96 & 5.09 \\
$1\times 10^{-3}$ & $3.23\times 10^{-3}$ & 4.51 & 7.44 \\
\bottomrule
\end{tabular}
\end{table}

\paragraph{Step-size sensitivity.} Table~\ref{tab:curvature_fd_sensitivity} repeats the probe at $\mathrm{Math}_o$ with $\eta=3\times10^{-4}$. The selected direction retains higher curvature than both random controls, while its absolute estimate changes from $3.23\times10^{-3}$ to $1.58\times10^{-3}$. The corresponding curvature ratios are $3.96$ and $5.09$, compared with $4.51$ and $7.44$ at $\eta=10^{-3}$. Both step sizes identify higher local sensitivity along the selected direction.

\end{document}